\documentclass[letterpaper, 10 pt, conference]{ieeeconf}  

\IEEEoverridecommandlockouts                              
\usepackage[utf8]{inputenc}
\usepackage[T1]{fontenc}
\usepackage{graphicx}
\usepackage{xcolor}
    \usepackage{caption}
    \usepackage{subcaption}
\usepackage{amsmath} 
\usepackage{amssymb}  
\usepackage{tikzscale}
\usepackage{pgfplots}
\usepackage[mode=buildnew]{standalone}
\usepgfplotslibrary{groupplots}
\usepackage{pgfplots, pgfplotstable}
\pgfplotsset{compat=1.17}
\newcommand{\putindex}[3]{\vtop{\hbox{\hspace{#3} $#1$}
            \hbox{\raise 6mm \hbox{$\scriptscriptstyle #2$}}}}

\newcommand{\gradx}[0]{\vtop{\hbox{\rm grad}
            \hbox{\raise 2.5mm \hbox{\rm \hspace{2mm} \footnotesize x}}}}

\newcommand{\grady}[0]{\vtop{\hbox{\rm grad}
            \hbox{\raise 2.5mm \hbox{\rm \hspace{2mm} \footnotesize y}}}}

\newcommand{\grad}[1]{\vtop{\hbox{\rm grad}
            \hbox{\raise 2.5mm \hbox{#1}}}}

\newcommand{\stz}{\rule{0mm}{2.3ex}}

\newcommand{\stzdown}{\rule[-1.2ex]{0mm}{3.5ex}}

\newcommand{\btb}{     \begin{tabbing}             }
\newcommand{\bte}{     \end{tabbing}               }

\usepackage{url}
\usepackage[labelformat=simple]{subcaption}

\definecolor{visible}{RGB}{35,139,69}
\definecolor{occluded}{RGB}{227,26,28}
\definecolor{cityscapes}{RGB}{247,125,0}
\definecolor{amcs}{RGB}{80,57,224}
\DeclareMathOperator*{\argmax}{argmax} 

\title{\LARGE \bf
Amodal Cityscapes: A New Dataset, its Generation,\\ and an Amodal Semantic Segmentation Challenge Baseline
}

\author{Jasmin Breitenstein$^{\ast}$ and Tim Fingscheidt$^{\ast}$
\thanks{$^{\ast}$Jasmin Breitenstein, and Tim Fingscheidt are with Institute for Communications Technology, Technische Universität Braunschweig, Schleinitz\-straße 22, 38106 Braunschweig, Germany, {\tt\small \{j.breitenstein, t.fingscheidt\}@tu-bs.de}}%
}

\begin{document}

\maketitle
\thispagestyle{empty}
\pagestyle{empty}

\begin{abstract}

Amodal perception terms the ability of humans to imagine the entire shapes of occluded objects. This gives humans an advantage to keep track of everything that is going on, especially in crowded situations. Typical perception functions, however, lack amodal perception abilities and are therefore at a disadvantage in situations with occlusions. Complex urban driving scenarios often experience many different types of occlusions and, therefore, amodal perception for automated vehicles is an important task to investigate. In this paper, we consider the task of amodal semantic segmentation and propose a generic way to generate datasets to train amodal semantic segmentation methods. We use this approach to generate an \textbf{\textit{amodal Cityscapes dataset}}. Moreover, we propose and evaluate a method as baseline on Amodal Cityscapes, showing its applicability for amodal semantic segmentation in automotive environment perception. We provide the means to re-generate this dataset on github\footnote{\textcolor{black}{\url{https://github.com/ifnspaml/AmodalCityscapes}}}.

\end{abstract}
\section{Introduction}

Visual perception is a crucial task for automated driving and its reliability is necessary for the development and distribution of intelligent vehicles. Visual perception itself comprises different tasks such as, e.g., object detection, instance and semantic segmentation. However, for reliable perception in traffic situations, a detection of corner cases is necessary. Corner cases have been defined as unpredictable events or appearances of relevant objects in locations relevant for the driving task \cite{Bolte2019b}. This definition has been refined into a systematization of corner cases in camera data \cite{Breitenstein2020}, and all available sensors in automated vehicles \cite{Heidecker2021b}. This refinement allows for specific development of methods treating certain corner case types, e.g., detecting an anomalous amount of objects in a scene \cite{Breitenstein2021}, or unknown objects \cite{Heidecker2021,Xia2020}.

\begin{figure}
\vspace*{.8em}
    \centering
    \includegraphics[width=0.46\textwidth]{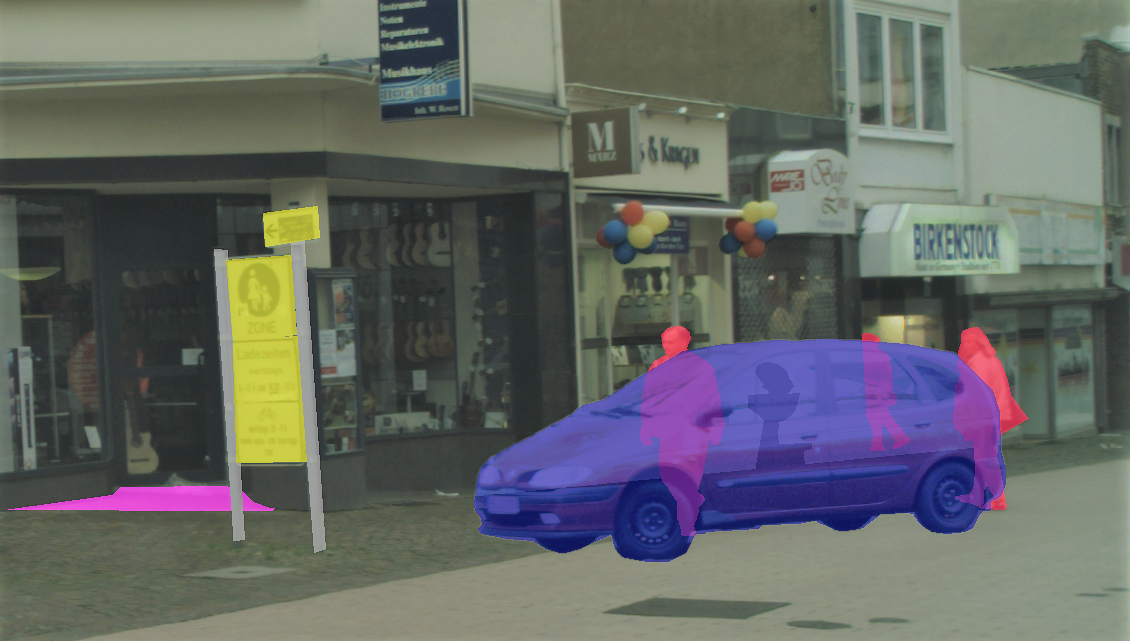}
    \caption{Example of a scene from the Amodal Cityscapes dataset with pedestrians (red) partially occluded by a car (blue). Amodal perception is the ability to conjecture the full shape of partially occluded objects as visualized in this image and provided as labels in the new dataset.}
    \label{fig:amodal-perception}
    \vspace*{-0.5cm}
\end{figure}

In this work, we consider the segmentation of occluded areas. On the one hand, occlusions are corner cases on their own since occluded objects are typically hard to detect \cite{Qi2019,Li2016}, on the other hand, occlusions can result in new corner cases because perception methods oftentimes have no awareness of what is happening behind the occlusions. Bogdoll et al. \cite{Bogdoll2021} provide an exemplary description of such a corner case.
Most existing visual perception methods focus on detecting or segmenting the visible objects and areas in camera data, however, humans are able to conjecture the \textit{in}visible components of a scene as well. For example, in Figure~\ref{fig:amodal-perception} in an example from the Amodal Cityscapes dataset, most visual perception methods would ideally recognize the visible parts of the persons (red, occludees) as a person, without taking into account that considerable portions of the persons are occluded behind the car (blue, occluder) in the foreground. \textit{Amodal} perception recognizes the (partly) occluded persons, while anticipating their entire shape, not just the visible parts. This yields essential visual cues for scene understanding, and thus forms a significant part of environment perception. 
One could argue that amodal perception should be performed after fusion in the occupancy grid/vector space  \cite{Erkent2018,Plachetka2021,Plebe2021}, or in some latent space non-interpretable to humans \cite{Ling2020,Yuting2021}. This work on amodal perception of simple camera data, however, comes with the clear advantage of (a) being human-readable, and (b) that already during multisensor fusion, the redundancy of multiple estimates of occluded objects may yield powerful confidence information.

When it comes to training data for amodal perception, there exist many large-scale automotive datasets, but we seldom come across amodal labels providing occlusion information. However, for semantic segmentation, we require not only occlusion information on object or scene level, but pixel-wise amodal labels. Considering the high costs and time required for labeling the visible parts of scenes, amodal labeling would be extremely challenging for large-scale datasets. Thus, typically, artificially augmented training data by inserting objects into the images to create amodal labeling is used to develop amodal visual perception methods \cite{Purkait2019,Ke2021}. Purkait et al. \cite{Purkait2019} augment their training data with completely randomly inserted instances for amodal semantic segmentation, which results in inserted objects in unusual image locations. Ke et al. \cite{Ke2021} train an amodal instance segmentation on datasets with general objects and amodal instance labels \cite{Zhu2017}. Qi et al. \cite{Qi2019} provide the KINS dataset with amodal instance labels on a subset of the KITTI dataset, however, without labels for the remaining (static) object classes required for semantic segmentation that describe a street scene in its entirety. Both \cite{Zhu2017, Qi2019} use their datasets to train amodal instance segmentations. However, for training of amodal semantic segmentation of automotive scenes, so far no dataset exists that provides \textit{both} amodal semantic segmentation labels and plausible occlusion locations.

Our contributions in this work are as follows: First, we describe a way to synthetically generate an amodal dataset on the basis of existing automotive datasets. Particularly, we provide code\footnote{\textcolor{black}{\url{https://github.com/ifnspaml/AmodalCityscapes}}} for automatic dataset generation to facilitate further research on amodal segmentation in automotive perception tasks. For this paper, we apply this approach to generate the \textit{Amodal Cityscapes} dataset based on the Cityscapes dataset \cite{Cordts2016}, which we characterize in detail. Second, we propose and evaluate a method which shall serve as baseline on our dataset, inviting others to take the challenge and to propose an improved method.
\section{Related Work}
\subsection{Datasets for Amodal Perception}

Amodal perception is of interest in many application fields of computer vision. Hence, datasets for amodal perception can be found in different fields, e.g., instance \cite{Zhu2017} and video instance segmentation \cite{Qi2021,Hu2019}, human recognition and de-occlusion \cite{Zhou2021}. The OVIS dataset \cite{Qi2021} provides instance masks for videos while additionally labeling the occlusion level of each instance. SAIL-VOS \cite{Hu2019} is a synthetic video instance segmentation dataset with amodal instance segmentation masks. The amodal human perception (AHP) dataset \cite{Zhou2021} is a large-scale video dataset which provides both modal and amodal masks focusing on humans. COCOA \cite{Zhu2017} contains both modal and amodal masks and is derived from the COCO dataset \cite{lin2014microsoft}. 

In the automotive setting, few datasets exist that provide amodal information. A common practice for pedestrian detection datasets is to provide both a visible and a full (possibly partly occluded) bounding box per object. This can be found in the CalTech Pedestrian \cite{Dollar2009}, CityPersons \cite{Zhang2017}, and KITTI \cite{Geiger2013} dataset. This KINS dataset \cite{Qi2019} extends this approach from KITTI to instance segmentation and provides amodal instance masks for selected scenes from the KITTI dataset. In this sense, it is to our knowledge the only automotive dataset providing pixelwise amodal labeling, but it is lacking semantic segmentation labels by focusing on instance segmentation, and thus labels for the remaining static classes relevant for environment perception of the underlying street scene are missing. The EuroCity Persons dataset \cite{Braun2019} provides annotated levels of occlusion per bounding box.
However, to train methods for amodal semantic segmentation, we require pixel-wise occlusion information. Thus, we aim to provide a method to automatically generate a dataset where pixel-wise amodal semantic segmentation labels can be obtained from the original data. This can be adapted to many datasets to simulate amodal labeling. Such a synthetic amodal dataset can be obtained via copy-paste of instances into the target images. This has previously been identified as strong data augmentation for visual perception method training \cite{Ghiasi2021, Dvornik2018,Remez2018, Dwibedi2017}, and also in amodal segmentation tasks \cite{Purkait2019,Ke2021}. In Section \ref{sec:dataset-generation}, we create such an amodal dataset from the Cityscapes dataset while maintaining plausible context for instances pasted into the target images.


\subsection{Amodal Semantic Segmentation}

The task of amodal segmentation has been defined by Zhu et al. \cite{Zhu2017} together with two baseline methods. The first directly predicts amodal masks from images, while the other expands upon a given visible segmentation mask \cite{Zhu2017}.
Purkait et al. \cite{Purkait2019} introduce an amodal semantic segmentation method, which extends the typical softmax layer to a groupwise softmax layer that is able to predict both visible and occluded semantic groups in images. 

The closely related task of amodal instance segmentation was first defined by Li and Malik \cite{Li2016} where they train a neural network to predict an amodal mask given the image and the corresponding visible mask. \texttt{AmodalMask} \cite{Zhu2017} is trained in a supervised manner and directly predicts the amodal mask.
Other approaches add a branch for occluded masks to the standard \texttt{Mask R-CNN} \cite{Follmann2019}, or combining the amodal and visible mask branches via multi-level coding \cite{Qi2019}.  

\section{Amodal Cityscapes Dataset Generation}
\label{sec:dataset-generation}

\begin{table}[t]
\vspace*{2pt}
    \centering
    \begin{tabular}{m{3cm}|m{1.25cm}|m{1.25cm}|m{1.25cm}}
    \hline
        \stz & \stz $\mathcal{D}_\text{amCS}^\text{train}$ &   \stz \stzdown $\mathcal{D}_\text{amCS}^\text{val}$ &  \stz $\mathcal{D}_\text{amCS}^\text{test}$ \\
         \hline \hline
       \stz \#images  & \stz $2900$ & \stz $75$ & \stz $500$ \\
       \stz \#ground truth masks & \stz $2900$ & \stz $75$ &\stz $500$ \\
       \stz \#target images &\stz $2900$ &\stz $75$ &\stz $500$ \\
       \stz \#source images &\stz $2900 -1$&\stz $75-1$ &\stz $500-1$\\
       \stz \#occluders available for pasting &\stz $36303-N$ & \stz $832-N$ & \stz $6438-N$\\
        \hline
    \end{tabular}
    \caption{Splits of the Amodal Cityscapes dataset $\mathcal{D}_\text{amCS}$.}
    \label{tab:amodal-cs-overview}
\end{table}

\begin{figure*}[t!]
\vspace*{2pt}
    \centering
    \begin{subfigure}[b]{0.3\textwidth}
                    \centering
     \includegraphics[width=\textwidth]{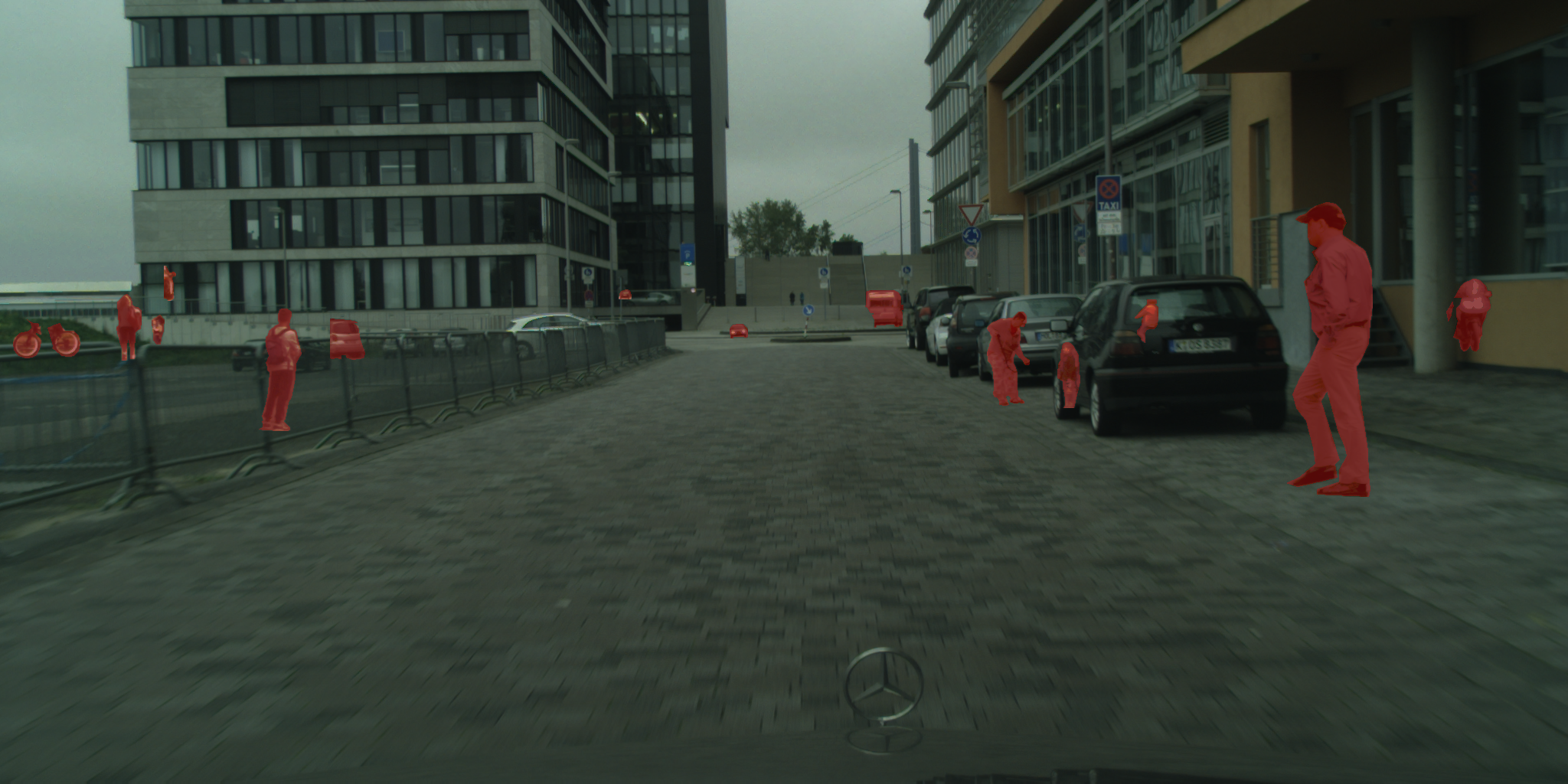}
    \caption{Example image $\mathbf{x}_t$ with inserted instances (in red for better visibility)}
    \label{subfig:example_image_amodalcs}
            \end{subfigure}
    \begin{subfigure}[b]{0.3\textwidth}
                    \centering
     \includegraphics[width=\textwidth]{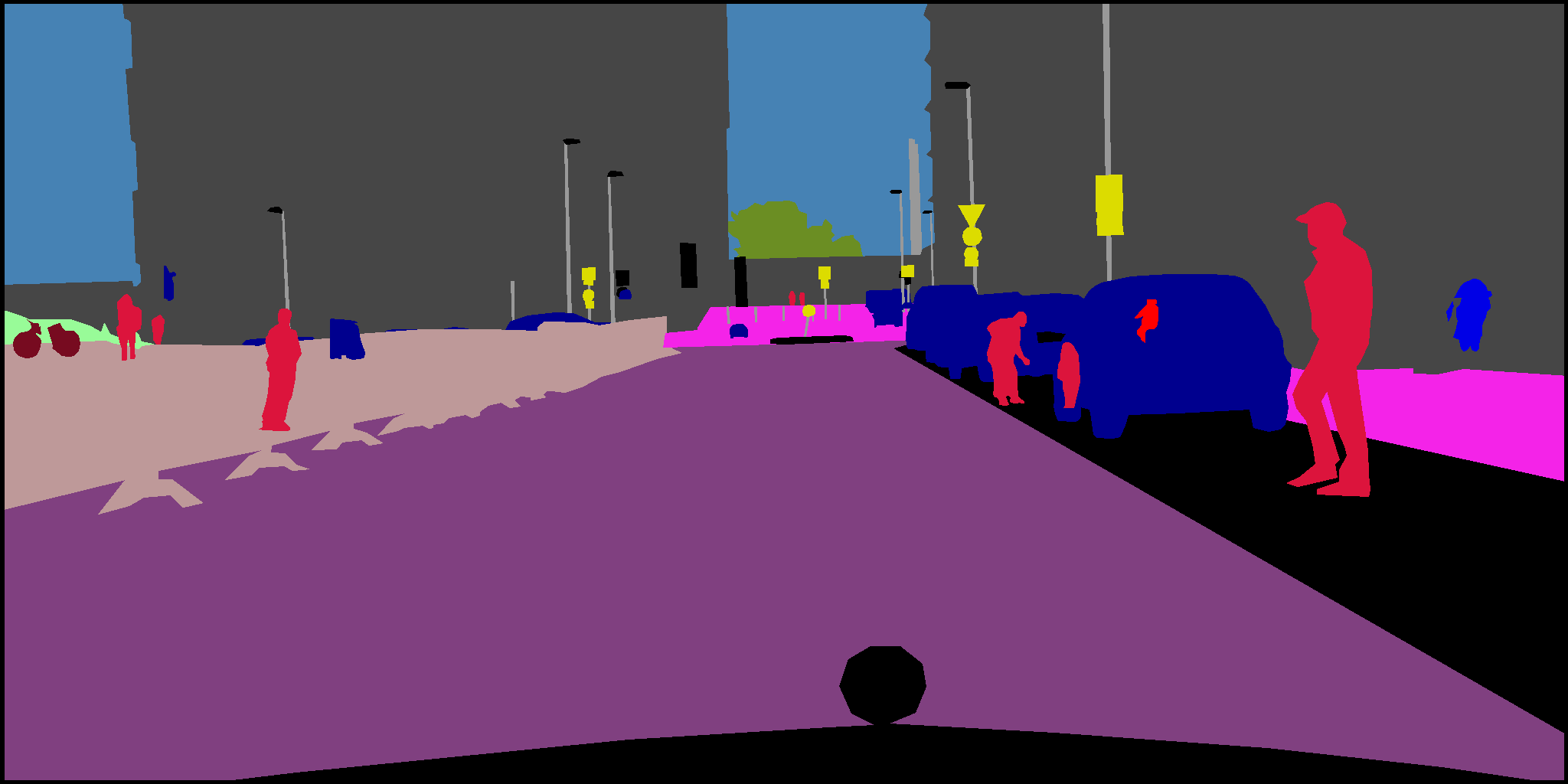}
    \caption{Ground truth semantic segmentation mask $\overline{\mathbf{m}}_{t,f=1}$}
    \label{subfig:example_visible_amodalcs}
            \end{subfigure}
                \begin{subfigure}[b]{0.3\textwidth}
                    \centering
     \includegraphics[width=\textwidth]{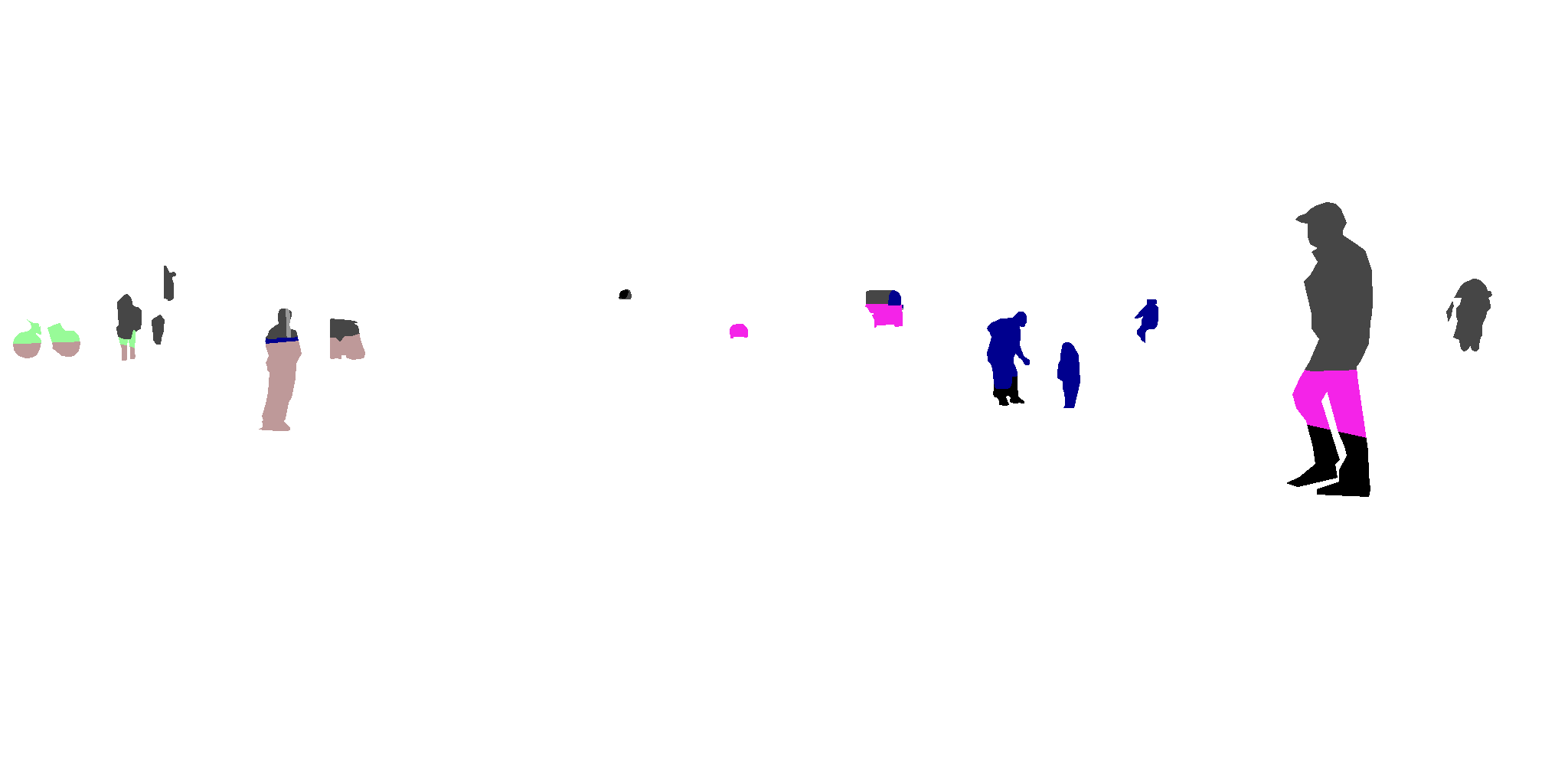}
    \caption{Ground truth amodal semantic segmentation mask $\overline{\mathbf{m}}_{t,f=2}$}
    \label{subfig:example_occluded_amodalcs}
            \end{subfigure}%
    \caption{Example image and segmentation masks from our Amodal Cityscapes training split.}
    \label{fig:example_amodalcs}
\end{figure*}

\textbf{Dataset splits.} In the following, we describe the generic way, in which we propose to generate a synthetic dataset for amodal semantic segmentation. In this paper, we take the Cityscapes dataset \cite{Cordts2016} and generate an amodal Cityscapes dataset. The Cityscapes dataset consists of $5000$ images of urban traffic scenes. The training set contains $2975$ images, the validation set contains $500$ images, and the test set contains $1525$ images. For both training and validation sets, ground truth labels in the form of semantic and instance segmentation masks are available. We use these two subsets to generate our amodal dataset. In the following, we use the notion \textit{target image} for the image into which we paste the occluders, i.e., occluding instances, and the notion \textit{source images} for the original images of those occluders.

Our resulting dataset consists of a training, validation and test split. As shown in Table \ref{tab:amodal-cs-overview}, for the training split $\mathcal{D}^\text{train}_\text{amCS}$, we select $2900$ images from the original Cityscapes training set as target images. The same $2900$ images serve as source images to extract the occluders. However, per target image, we select only occluders from other images, resulting in a total of $2900-1=2899$ source images with $36303-N$ occluders available for pasting into this target image. Here, $36303$ is the total number of instances from all the source images, and $N$ denotes the number of instances of the current target image. For the validation split $\mathcal{D}^\text{val}_\text{amCS}$, we take the remaining $75$ images from the original Cityscapes training set as target images. Again, the same images serve as source images for the occluders, resulting in $75-1=74$ source images with $832-N$ occluders per target image. For the test split $\mathcal{D}^\text{test}_\text{amCS}$, we follow common practice and use the validation set of the Cityscapes dataset. Thus, we have $500$ target images, with up to $500-1$ source images and $6438-N$ occluders available for pasting into this target image.

\begin{figure}
    \centering
    \begin{subfigure}[b]{0.23\textwidth}
                    \centering
     \includegraphics[width=\textwidth]{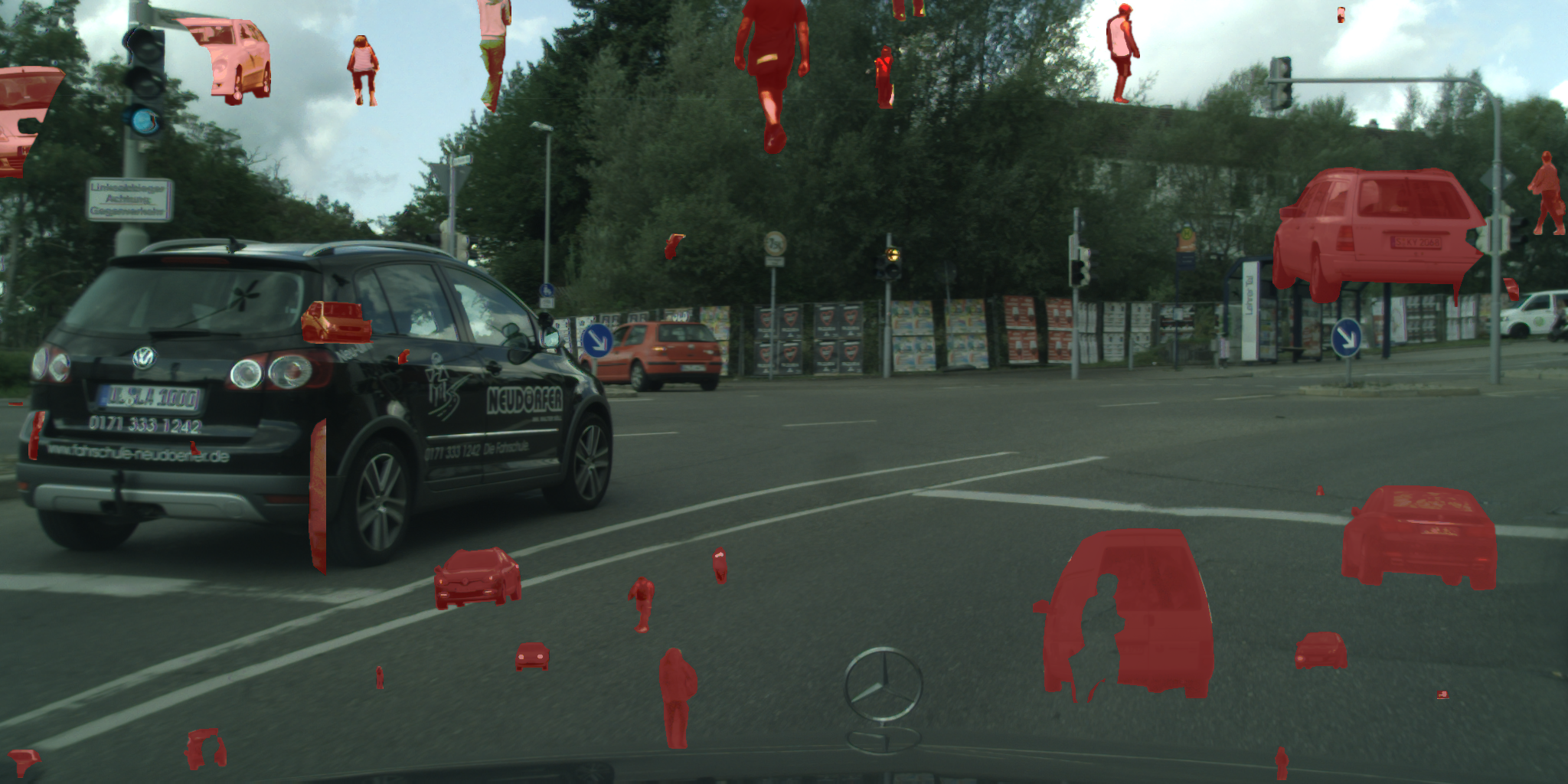}
            \end{subfigure}%
            \hspace{1pt}
    \begin{subfigure}[b]{0.23\textwidth}
                    \centering
     \includegraphics[width=\textwidth]{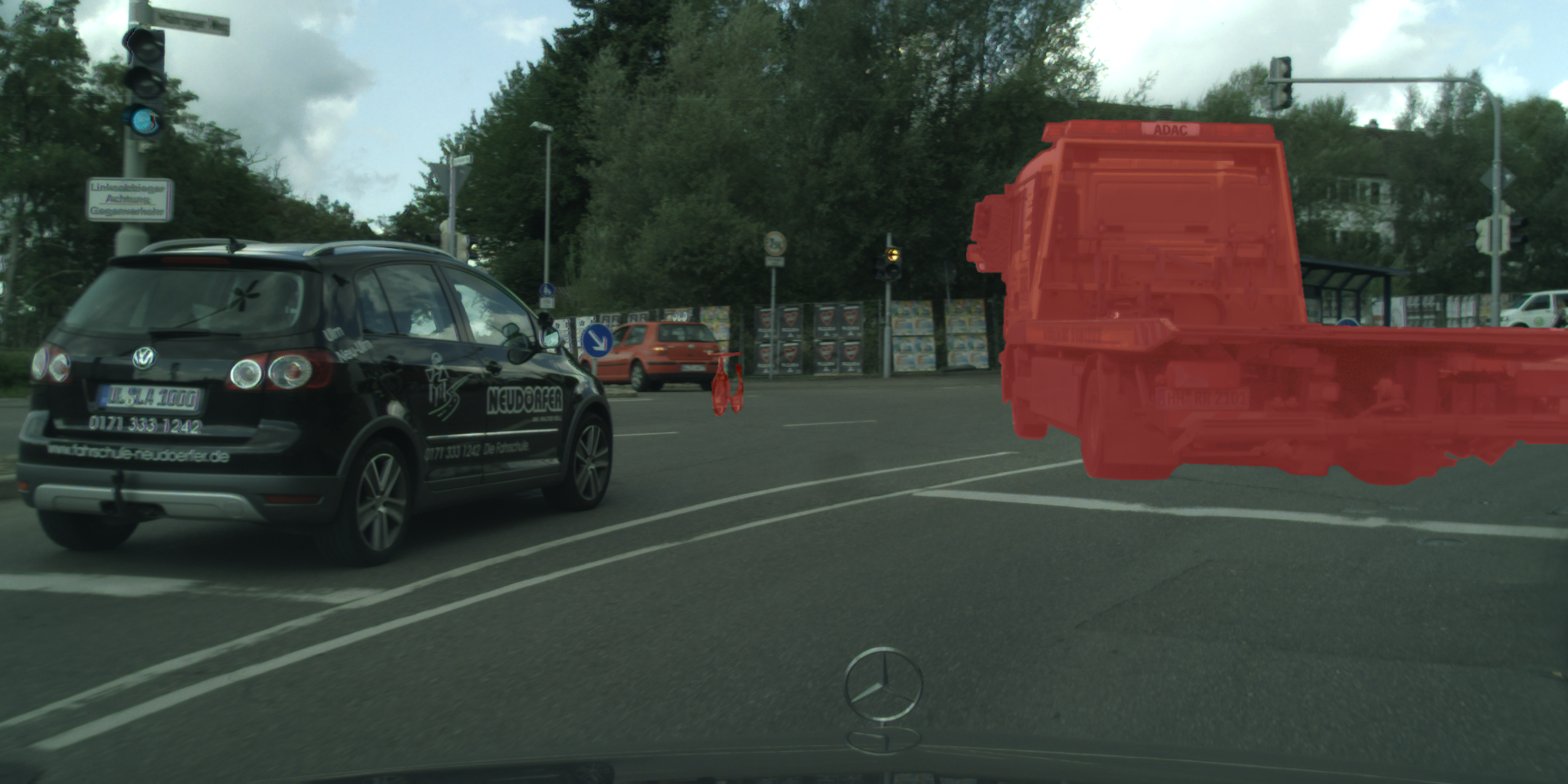}
            \end{subfigure}%
    \caption{Example image with occluders inserted randomly (left) and based on the original location as proposed (right). Occluders are shaded in red for better visibility.}
    \label{fig:plausiblelocation}
\end{figure}

\textbf{Dataset generation.} In a first step, we use the ground truth instance segmentation masks, to extract the available instances from the source images $\tilde{\mathbf{x}} \in \mathbb{I}^{H \times W \times C} $, with $  \mathbb{I}= \lbrack 0,1 \rbrack$, and height $H$, width $W$, and number of channels $C=3$, respectively. We extract the annotated instances, $\mathbf{a}_{n} = (a_{n,i}) \in \lbrace 0,1 \rbrace^{H \times W}$, $n \in \mathcal{N} = \lbrace 1,\ldots, N \rbrace$, and $i \in \mathcal{I}= \lbrace 1,\ldots, H \cdot W \rbrace$. Here, $N$ denotes the number of instances in $\tilde{\mathbf{x}}$, with $s_n \in \mathcal{S} = \lbrace 1, \ldots, S \rbrace$ being the respective semantic class of instance $n$, saving both the RGB instance image and the corresponding binary mask $\mathbf{a}_n$, along with its class $s_n$. We filter out too small or too occluded instances by choosing a minimum size of $10 \times 20$ pixels.

Then, we use the extracted instance masks to generate our copy-paste amodal dataset. For each target image $\hat{\mathbf{x}}$ of the respective dataset split, we select a number of instances to be pasted into the image as occluder and, for each instance, we carefully decide on a location in the image to paste the instance. Both occluders and occludees are chosen from the Cityscapes dataset, hence they can be partially occluded. By fixing a minimum occluder size, we filter out heavy partial occlusions. However, from our experiments we find that the intrinsic partial occlusions do not hinder the method in performing amodal semantic segmentation.

The \textit{number of occluders} $\hat{N}$ for a target image is selected to fit a randomly chosen occlusion ratio $P_\mathrm{o} \in \lbrack 0,0.1 \rbrack$, determining the percentage of occluded pixels in the resulting image. The occlusion ratio is reached, once the insertion of an object just exceeds the drawn ratio for that target image.
The \textit{location for pasting an occluder} is chosen according to its original location in the source image. A location per instance is prescribed by the lower left corner of its bounding box, i.e., a pixel location $(h,w)$ with $h$ the vertical, and $w$ the horizontal index. Alternatively, pixel index  $i \in \mathcal{I}$ could be chosen. While we allow $w$ to be chosen randomly over the entire image width as in \cite{Purkait2019}, the vertical indices $h$ of the occluder remain fixed (different to \cite{Purkait2019}!). This ensures object location plausibility in the resulting images. Figure~\ref{fig:plausiblelocation} shows the effect of this method: While randomly chosen locations result in, e.g., cars occluding parts of the sky and less reasonable instance sizes, the strategy based on the original vertical position results in more plausibly placed and sized occluders. While more complex placement methods for the source instances are possible, the simple heuristic of fixing the vertical position delivers valid results. While we can still observe some less plausible placements, these are due to actual instance positions in the original dataset. For example, a person instance placed adjacent to the sky results from source images, e.g., with a high street horizon due to tilts of the ego-vehicle, or bridges. Hence, more complex placement methods restricting the positioning of pedestrians to, e.g., the sidewalk, do not necessarily depict reality better, but instead might make the perception method more vulnerable to corner cases.

We overwrite the RGB image with the new instance while in the semantic segmentation mask we keep both the new visible labels and the previous, now occluded, labels. To ensure that the amodal semantic segmentation method does not learn to detect sharp edges, we use a Gaussian filter of size $5 \times 5$ to blend the edges of the inserted occluders with the target image as is done in \cite{Dwibedi2017}.

\setcounter{figure}{4}
    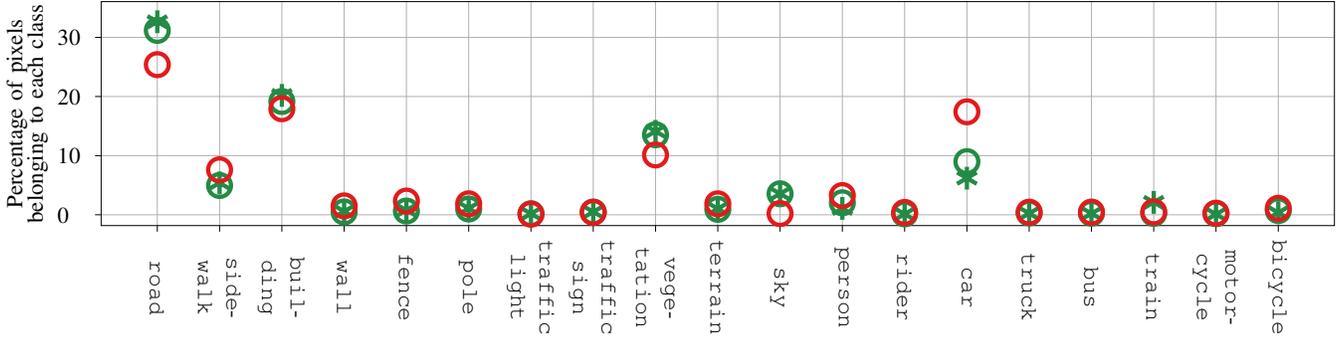
\begin{figure*}[ht!]
    \vspace*{2pt}
                    \centering
\begin{tikzpicture}
\definecolor{color0}{RGB}{35,139,69}
\definecolor{color1}{RGB}{227,26,28}
\begin{axis}[
tick align=outside,
tick pos=left,
x grid style={white!69.0196078431373!black},
xmajorgrids,
width=1.8\textwidth,
height = 0.4\textwidth,
xmin=-0.901774193548387, xmax=18.9017741935484,
xtick style={color=black},
xtick={0,1,2,3,4,5,6,7,8,9,10,11,12,13,14,15,16,17,18},
xticklabel style={font=\LARGE, text width=2.5cm, align=center,rotate=-90},
xticklabels={\texttt{road},\texttt{side}-\texttt{walk},\texttt{buil}-\texttt{ding},\texttt{wall},\texttt{fence},\texttt{pole},\texttt{traffic} \texttt{light},\texttt{traffic} \texttt{sign}, \texttt{vege}-\texttt{tation}, \texttt{terrain}, \texttt{sky}, \texttt{person}, \texttt{rider}, \texttt{car},\texttt{truck},\texttt{bus},\texttt{train},\texttt{motor}-\texttt{cycle},\texttt{bicycle}},
y grid style={white!69.0196078431373!black},
ymajorgrids,
ymin=-0.0182922023410576, ymax=0.3605565283979908,
yticklabel style={font=\LARGE, text width=0.8cm},
ytick style={color=black},
yticklabels={0,0,10,20,30},
ylabel style={align=center},
ylabel={\LARGE Percentage of pixels\\ \LARGE belonging to each class},
]
\addplot [only marks, mark options={scale=4, line width=3pt}, mark=o, draw=color0, fill=color0, colormap/viridis]
table{%
x                      y
0 0.31178924
1 0.04949556
2 0.19159702
3 0.0047737
4 0.00630149
5 0.00979002
6 0.00177026
7 0.00465982
8 0.13529849
9 0.00921237
10 0.03546953
11 0.02026255
12 0.00151702
13 0.08980653
14 0.0030628
15 0.00272663
16 0.00243789
17 0.00188473
18 0.00711001
};
\addplot [only marks, mark options={scale=4, line width=3pt}, mark=asterisk, draw=color0, fill=color0, colormap/viridis]
table{%
x                      y
0 0.32637906
1 0.05386518
2 0.2019027
3 0.00563657
4 0.00763402
5 0.01093374
6 0.00182922
7 0.00489837
8 0.14111377
9 0.01027697
10 0.03558883
11 0.01084613
12 0.00119019
13 0.06192826
14 0.00233634
15 0.00209501
16 0.020989
17 0.0008769
18 0.00364826
};

\addplot [only marks, mark options={scale=4, line width=3pt}, mark=o, draw=color1, fill=color1, colormap/viridis]
table{%
x                      y
0 0.25388085
1 0.07603683
2 0.1793315
3 0.01501505
4 0.02318776
5 0.01852575
6 0.00102599
7 0.00415113
8 0.10119317
9 0.01852532
10 0.00207601
11 0.03249166
12 0.00397344
13 0.17417259
14 0.00439421
15 0.00453251
16 0.00482035
17 0.0026879
18 0.01095913
};
\end{axis}

\end{tikzpicture}
    \caption{Percentage of pixels belonging to each semantic class on Cityscapes ($\ast$) and Amodal Cityscapes ($\circ$) for both \textcolor{visible}{visible (green)} and \textcolor{occluded}{occluded (red)} areas. }
    \label{fig:cs-amcs-classfrequencies}
        \end{figure*}

\setcounter{figure}{3}
\begin{figure}[h]
    \centering
    \begin{subfigure}[b]{0.23\textwidth}
                    \centering
        \includegraphics[width=\textwidth]{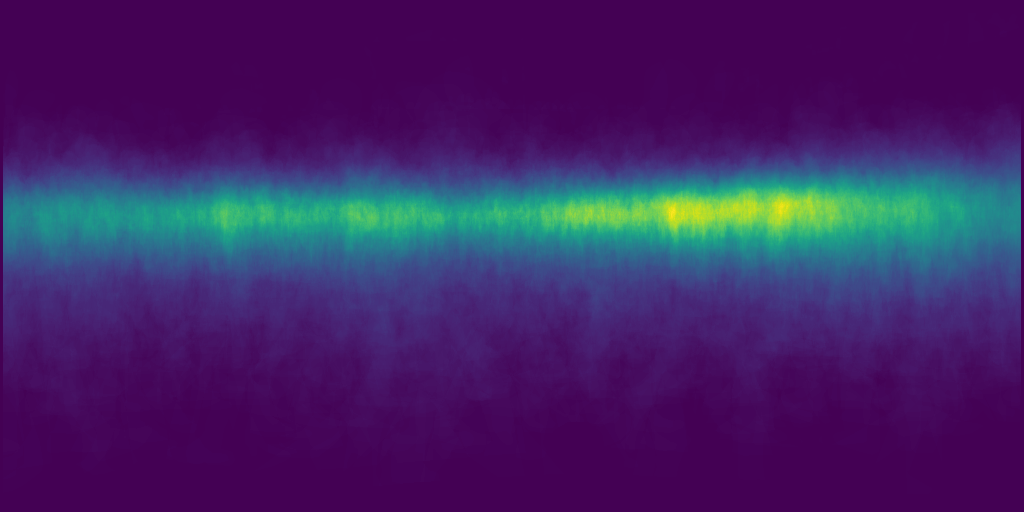}
    \end{subfigure}
        \begin{subfigure}[b]{0.23\textwidth}
                    \centering
        \includegraphics[width=\textwidth]{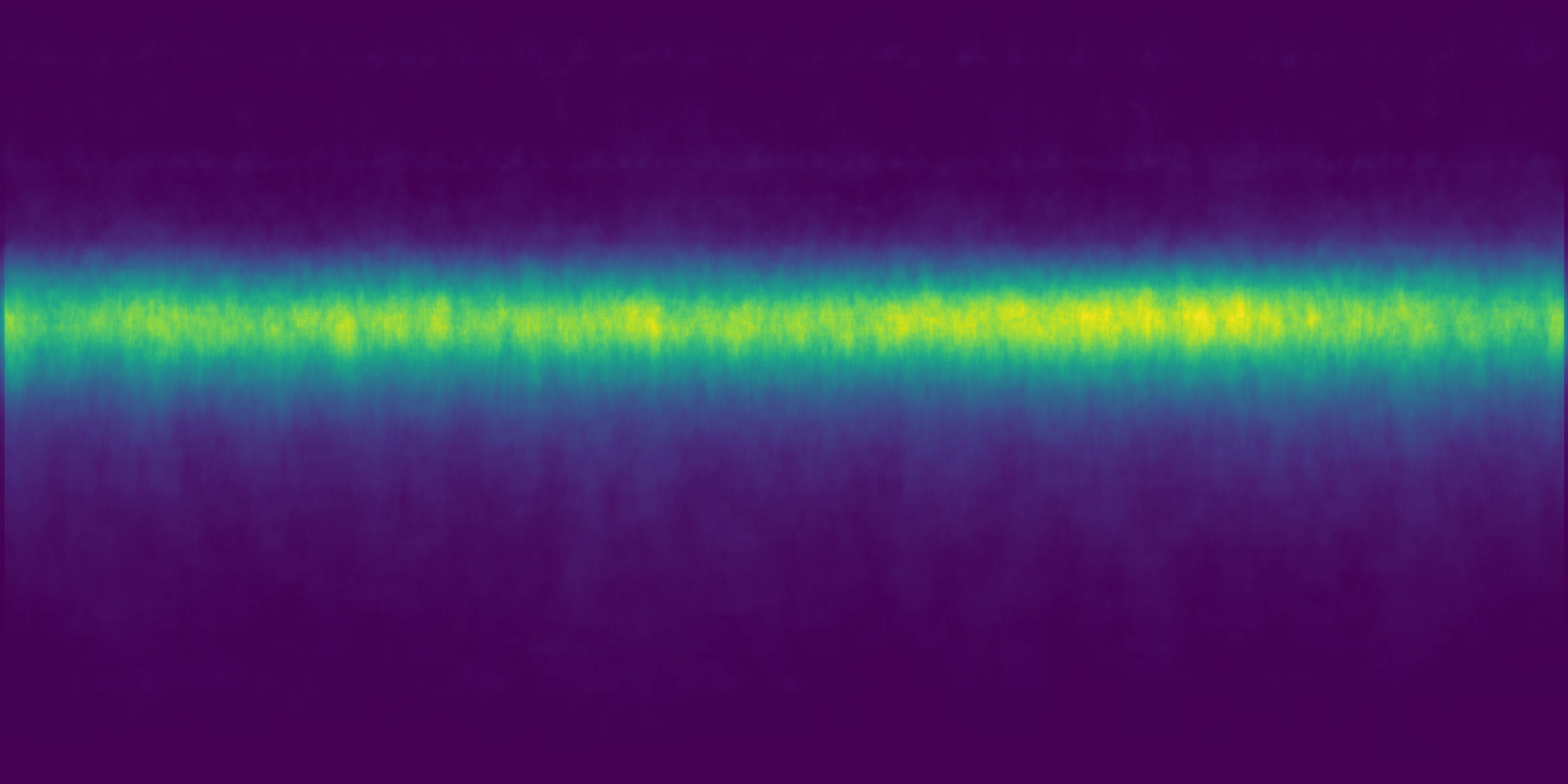}
    \end{subfigure}
    \caption{Location priors for class \texttt{person} on the training splits of Cityscapes ($\mathcal{D}_\text{CS}^\text{train}$, left) and \textcolor{black}{Amodal Cityscapes ($\mathcal{D}_\text{amCS}^\text{train}$, right)}. As intended, they appear to be similar.}
    \label{fig:locationpriors}
\end{figure}

Each resulting data split can be formally described as images $\mathbf{x}_1^T = \lbrace \mathbf{x}_t \rbrace$, with $T = \lbrace 2900,75,500 \rbrace$. 
An example image can be seen in Figure~\ref{subfig:example_image_amodalcs}. For each image, there is a corresponding amodal ground truth semantic segmentation mask $\overline{\mathbf{m}}_{t}=\left( \overline{\mathbf{m}}_{t,i} \right) = \left( \overline{m}_{t,i,f} \right)  \in \mathcal{S}^{H \times W \times F}$, with $\mathcal{S}$ being the set of semantic classes $\mathcal{S} = \lbrace 1 , \ldots , S \rbrace$, and $S=19$ for Cityscapes. We use the channel size $F=2$ in this case to encode the visible ($f=1$) and occluded ($f=2$) pixels in the image, i.e., $\overline{m}_{t,i,1}=s_1 \in \mathcal{S}$, see Figure~\ref{subfig:example_visible_amodalcs}, and $\overline{m}_{t,i,2}=s_2 \in \mathcal{S}$, see Figure~\ref{subfig:example_occluded_amodalcs}, meaning that in pixel position $i$ class $s_1$ is visible and class $s_2$ is occluded. For this work, we chose $F=2$, since most amodal segmentation methods only aim at two layers, foreground and background. To ensure $F=2$, we took care that occluders are not occluding each other.

\textbf{Dataset statistics.} Finally, we compare the Cityscapes dataset ($\mathcal{D}_\text{CS}$) with the Amodal Cityscapes ($\mathcal{D}_\text{amCS}$) dataset. Most importantly, to keep plausibility, we show that the distribution of dynamic objects, i.e., all instances listed in Table \ref{tab:comparison-amcs}, does not change. Figure~\ref{fig:locationpriors} shows the location priors for the class \texttt{person} on the training splits of both datasets. We observe no significant distribution changes.

Since per definition of Amodal Cityscapes $\mathcal{D}_\text{amCS}$, we insert dynamic objects, clearly the total number of dynamic objects and the number of dynamic objects per image increase, see Table \ref{tab:comparison-amcs} for the training dataset splits. However, while the total number of all dynamic instances increases due to the insertions, Table \ref{tab:comparison-amcs} shows that the percentage of an instance class from the total number of instances in this case for the training dataset splits does not change extremely, e.g., \textcolor{black}{in Cityscapes $\mathcal{D}^\text{train}_\text{CS}$, \texttt{person} instances contribute 34.4\% to the total instances, while in Amodal Cityscapes $\mathcal{D}^\text{train}_\text{amCS}$ they are 46.5\% of the total instances, which is the largest discrepancy for the instance classes}. 
In Figure~\ref{fig:cs-amcs-classfrequencies}, we see that the relative ordering of the visible semantic classes in terms of their appearance frequency remains very similar. 
Figure~\ref{fig:cs-amcs-classfrequencies} shows additionally for each semantic class the percentage of occluded pixels that belong to that class. Compared to the visible areas, we see that the distribution of occluded pixels to each class is very similar to the general distribution of visible pixels on the Cityscapes dataset.

\begin{table}[t!]
    \centering
    \begin{tabular}{m{3.75cm}|m{1.78cm}|m{1.78cm}}
    \hline
         \stz & \stz \stzdown $\mathcal{D}_\text{CS}^\text{train}$ & \stz  \stzdown $\mathcal{D}_\text{amCS}^\text{train}$ \\
         \hline \hline
         \stz \#images \stz & 2,975 & 2,900\\
         occlusion ratio $P_\mathrm{o}$& 0 & 0.06 \\
         \#dynamic objects & \textcolor{black}{50,782}& \textcolor{black}{100,395}\\
         \#\texttt{person} instances \newline (\% of instances) & \textcolor{black}{17,466 (34.4\%)}& \textcolor{black}{66,748 (46.5\%)}\\
         \#\texttt{car} instances (\% of instances) & \textcolor{black}{26,296 (51.8\%)}& \textcolor{black}{59,660 (41.6\%)}\\
         \#\texttt{rider} instances \newline (\% of instances) & \textcolor{black}{1,707 (3.4\%)}& \textcolor{black}{3,718 (2.6\%)}\\
         \#\texttt{motorcycle} instances \newline (\% of instances) & \textcolor{black}{720 (1.4\%)}& \textcolor{black}{1,797 (1.3\%)}\\
         \#\texttt{train} instances \newline (\% of instances) & \textcolor{black}{165 (0.3\%)}& \textcolor{black}{284 (0.2\%)}\\
         \#\texttt{truck} \newline instances (\% of instances) & \textcolor{black}{467 (0.9\%)}& \textcolor{black}{1102 (0.8\%)}\\
         \#\texttt{bus} instances (\% of instances) & \textcolor{black}{373 (0.7\%)}& \textcolor{black}{679 (0.5\%)}\\
         \#\texttt{bicycle} instances \newline (\% of instances) & \textcolor{black}{3,588 (7.1\%)}& \textcolor{black}{9,545 (6.7\%)}\\
         \hline
    \end{tabular}
    \caption{Some statistics on training splits of (Amodal) Cityscapes with occurrences of dynamic object classes.}
    \label{tab:comparison-amcs}
\end{table}


\section{New Amodal Semantic Segmentation Baseline Method}

\begin{table}[t]
    \centering
    \begin{tabular}{p{1.9cm}|l|l|p{3.8cm}}
    \hline
        Group name & \stz \stzdown $k$ & \stz \stzdown $g_k + 1$ & \stz Semantic classes \\
         \hline \hline
        \stz Static & \stz $0$ & \stz $8+1$ & \stz \texttt{road, sidewalk, building, wall, sky, terrain, fence, vegetation}, absence \\
        Traffic objects & $1$ & $3+1$ & \texttt{traffic signs, traffic light, pole}, absence \\
        Person-like & $2$ & $2+1$ & \texttt{person, rider}, absence  \\
        Vehicle-like & $3$ & $6+1$ &\texttt{car, truck, bus, train, bicycle, motorcycle}, absence  \\
         \hline
    \end{tabular}
    \caption{Grouping of the semantic classes in the amodal Cityscapes dataset with $K=4$.}
    \label{tab:semantic-grouping}
\end{table}

We base our amodal semantic segmentation method that we propose as dataset (or challenge) baseline on the \texttt{ERFNet} \cite{Romera2018}, and follow the proposed approach of Purkait et al. \cite{Purkait2019}
for amodal semantic segmentation. They employ an adaptation of the standard softmax layer to predict also invisible semantic groups. This approach is based on grouping the $S$ semantic classes into $K$ groups $\mathcal{G}_k$, $k \in \mathcal{K}= \lbrace 0, \ldots, K-1 \rbrace$, each of size $g_k = | \mathcal{G}_k|$. Additionally, per group, also the absence of the group is encoded, resulting in $g_k+1$ options per group, for details see Table \ref{tab:semantic-grouping}.

While we follow the grouping from Purkait et al. \cite{Purkait2019} for the definition of the \textit{static} ($\mathcal{G}_0$) and \textit{traffic object} ($\mathcal{G}_1$) group, we split the remaining semantic classes in the \textit{person-like} ($\mathcal{G}_2$) and the \textit{vehicle-like group} ($\mathcal{G}_3$), since occluded pedestrians are of special interest for safe automated driving. This setting is denoted as $K=4$ groups, while for $K=3$, the setting proposed in \cite{Purkait2019}, the person-like and vehicle-like group are fused. In this setting for amodal semantic segmentation, semantic classes belonging to the same group cannot occlude each other. In semantic segmentation we do not distinguish, e.g., between two different persons, hence one person occluding the other is incoherent with that definition, and irrelevant to the driving task. Thus, this restriction is a valid approach for amodal semantic segmentation.

\setcounter{figure}{5}
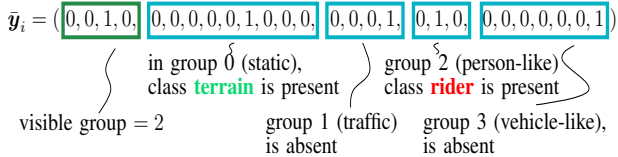
\begin{figure}
\vspace*{2pt}
    \centering
     \resizebox{8.5cm}{2.1cm}{\begin{tikzpicture}[node distance = 2cm]
\usetikzlibrary{shapes.geometric, arrows.meta, calc,backgrounds}
\definecolor{color0}{RGB}{35,139,69}
\definecolor{color1}{RGB}{1,179,196}
\definecolor{person}{RGB}{220,20,60}
\definecolor{road}{RGB}{128, 64,128}
\definecolor{rider}{RGB}{255,0,0}
\definecolor{building}{RGB}{70, 70, 70}
\definecolor{terrain}{RGB}{7, 217, 112}
\node at (-3,1) {\large $\bar{\boldsymbol{y}}_i = ( \; 0,0,1,0,\; \; 0,0,0,0,0,1,0,0,0, \;\; 0,0,0,1, \;\; 0,1,0,  \; \; 0,0,0,0,0,0,1 \;) $
};

\draw[color0, line width=2pt]  (-8.25,1.3) rectangle (-6.65,0.7);
\draw[color1, line width=2pt]  (-6.45,1.3) rectangle (-2.9,0.7);
\draw[color1, line width=2pt]   (-2.7,1.3) rectangle (-1.1,0.7);
\draw[color1, line width=2pt]   (-0.88,1.3) rectangle (0.3,0.7);
\draw[color1, line width=2pt]   (0.55,1.3) rectangle (3.2,0.7);

\draw[line width = 0.5pt] plot[smooth, tension=.7] coordinates {(-7.15,0.5)
(-7.35,0.15) (-7.25,-0.25) (-7.5,-0.65)};
\node at (-7.65,-0.85) {\large visible group $= 2$};

\draw[line width = 0.5pt] plot[smooth, tension=.7] coordinates {(2.5,0.55)
(2.25,0.3) (2.8,-0.25) (1.7,-0.5)};
\node[align=left] at (-4.45,0) {\large in group 0 (static), \\\large class \textbf{\textcolor{terrain}{terrain}} is present};

\draw[line width = 0.5pt] plot[smooth, tension=.7] coordinates {(-1.95,0.6)
(-1.8,0.1) (-2.15,-0.25) (-2.2,-0.55)};
\node[align=left] at (-2.6,-1) {\large group 1 (traffic)\\\large is absent};

\draw[line width = 0.5pt] plot[smooth, tension=.7] coordinates {(-0.4,0.4)
(-0.35,0.45) (-0.45,0.55) (-0.35,0.6)};
\node[align=left] at (0.35,0) {\large group 2 (person-like)\\ \large class \textbf{\textcolor{rider}{rider}} is present};

\draw[line width = 0.5pt] plot[smooth, tension=.7] coordinates {(-4.65,0.6)
(-4.75,0.55) (-4.7,0.45) (-4.75,0.4)};
\node[align=left] at (1.25,-1) {\large group 3 (vehicle-like), \\\large is absent};
\end{tikzpicture}}
    \caption{Example encoding of the amodal ground truth in a pixel with visible rider and occluded terrain for groupwise segmentation. Only one group can be visible. Here, $K=4$, meaning that including the static group, up to three objects from the invisible three groups can in principle be encoded.}
    \label{fig:example-encoding}
\end{figure}

This grouping requires us to also encode the ground truth mask $\overline{\mathbf{m}}_{t} \in \mathcal{S}^{H \times W \times F}$ in this groupwise setting: The first channel, i.e., $\overline{\mathbf{m}}_{t,f=1}$ provides the visible semantic class $s_1 \in \mathcal{S}$, and hence group $k_1 \in \mathcal{K}$ for each pixel, the second, amodal channel $\overline{\mathbf{m}}_{t,f=2}$ provides the occluded semantic class $s_2 \in \mathcal{S}$ and corresponding group $k_2 \in \mathcal{K}$. This encoding can be performed as groupwise one-hot encoding: Figure~\ref{fig:example-encoding} shows the encoding $\overline{\mathbf{y}}_{t,i}$ for a pixel $i$ with visible class \texttt{rider} and occluded class \texttt{terrain} based on the $K=4$ grouping as provided in Table \ref{tab:semantic-grouping}. More formally, for $K$ groups the ground truth mask $\overline{\mathbf{m}}_{t}$ is encoded as
\vspace*{-5pt} \begin{equation}
    \overline{\mathbf{y}}_t = \left( \overline{\mathbf{y}}_{t,i} \right) \in \lbrace 0,1 \rbrace^{H \times W \times \left( K + \sum\limits_{k \in \mathcal{K}} (g_k +1 )\right)},
\end{equation}
which is a vector of length $\left( K + \sum\limits_{k\in \mathcal{K}} (g_k +1) \right)$ per pixel, whereby the $K$ first elements encode the visible group (green rectangle in Figure~\ref{fig:example-encoding}). The blue rectangles in Figure~\ref{fig:example-encoding} highlight the parts of the vector corresponding to the $K$ groups and encode the particular class within each group, and also the presence of the group as a whole.

In general, in amodal semantic segmentation, the network takes as input a (normalized) image $\mathbf{x}_t \in \mathbb{I}^{H \times W \times C}$. In the standard softmax layer, one obtains per input image $\mathbf{x}_t$ an output $\mathbf{y}_t = (\mathbf{y}_{t,i}) \in \lbrack 0, 1 \rbrack^{H \times W \times S}$. We replace the softmax layer of the \texttt{ERFNet} with the groupwise softmax final layer proposed in \cite{Purkait2019}, which yields now an output tensor
\begin{equation}
    \mathbf{y}_t = (\mathbf{y}_{t,i}) \in \lbrack 0, 1 \rbrack^{H \times W \times \left( K + \sum\limits_{k \in \mathcal{K} } (g_k+1) \right) } .
    \label{eq:groupwise-softmax-outputs}
\end{equation}
For the following, we denote $\mathbf{y}_{t,i} = \left( \mathbf{p},\mathbf{q}_0, \ldots,\mathbf{q}_{K-1} \right)$ with $\mathbf{p} \in \lbrack 0,1 \rbrack^K$ and $\mathbf{q}_k \in \lbrack 0,1 \rbrack^{g_k+1}$, to specify the components of the factors in \eqref{eq:groupwise-softmax-outputs}, as shown by the rectangles in Figure~\ref{fig:example-encoding}.

Given such an output tensor $\mathbf{y}_{t,i}$, we can still recover the (visible) semantic segmentation mask: Defining functions $f_k(\cdot)$ , $k \in \mathcal{K}$, to project the respective indices of $\mathbf{q}_k$ to the corresponding semantic class $s$, we obtain the visible semantic class for pixel $i$ as 
\begin{equation}
    m_{t,i,f=1} = f_{k'} \left( \text{argmax}(\mathbf{q}_{k'} ) \right) \in \mathcal{S},
     \label{eq:mtif1}
\end{equation}

\noindent with the group index of the visible group $k'$ = $\text{argmax}(\mathbf{p})$. In Figure~\ref{fig:example-encoding}, we would obtain $\mathbf{m}_{t,i,f=1}=12$ (class \texttt{rider} in Cityscapes), and $k'=2$ as the visible group index. The group index $k''$ corresponding to the second largest value of $\mathbf{p}$ gives the occluded semantic class per pixel as
\begin{equation}
    m_{t,i,f=2} = f_{k''} \left( \text{argmax}(\mathbf{q}_{k''} ) \right) \in \mathcal{S}.
    \label{eq:mtif2}
\end{equation}
In Figure~\ref{fig:example-encoding}, we would obtain $\mathbf{m}_{t,i,f=2}=9$ (class \texttt{terrain} in Cityscapes), and $k''=0$ as the occluded group index during inference for a not one-hot network output $\mathbf{y}_{t,i}$. \textit{It is important to note that in the configuration of \eqref{eq:mtif1} and \eqref{eq:mtif2}, only one group may contain a visible object and a different group may contain an occluded object while the remaining two groups are always absent. With this choice, we follow \cite{Purkait2019}.} Still, according to Figure~\ref{fig:example-encoding}, the network always delivers a visible group with a respective visible class, and for $K=4$ three invisible groups with a most likely occluded class in each group. While Amodal Cityscapes allows objects from the same class to occlude each other, the method does not treat those occlusions since in the setting of semantic segmentation the presence of a certain semantic class in each pixel provides sufficient knowledge for the driving task as we do not distinguish between different instances. Instead, during inference, the method directly predicts another group with an occluded class in the amodal semantic segmentation $m_{t,i,f=2}$ that is considered to be behind the semantic classes from the visible group.

Next to the amodal mask prediction, the method allows to obtain the groups and corresponding semantic classes present in each pixel, due to the separate treatment of the $K$ groups in the encoding \eqref{eq:groupwise-softmax-outputs}. Hence, we predict per pixel position $i$ whether a group is absent in this position, or, if not, which semantic class $s_k \in \mathcal{S}_k$ of the group $\mathcal{G}_k$, $k \in \mathcal{K}$ is present in this pixel. We denote as $\mathcal{S}_k \subset \mathcal{S}$ the set of semantic classes contained in group $\mathcal{G}_k$, and as $q_k(s_k)$ the entry of $\mathbf{q}_k = (q_k(s_k))$ corresponding to semantic class $s_k$. Thus, we additionally obtain a mask of present semantic classes per pixel position $i$ \textit{for each group} $\mathcal{G}_k$, $k \in \mathcal{K}$ by
\begin{equation}
   \argmax(\mathbf{q}_k) := \argmax_{s_k \in \mathcal{S}_k} (q_k(s_k)).
\end{equation}

 \begin{figure*}[ht!]
 \vspace*{2pt}
    \centering
    \begin{subfigure}[b]{0.19\textwidth}
                    \centering
     \includegraphics[width=\textwidth]{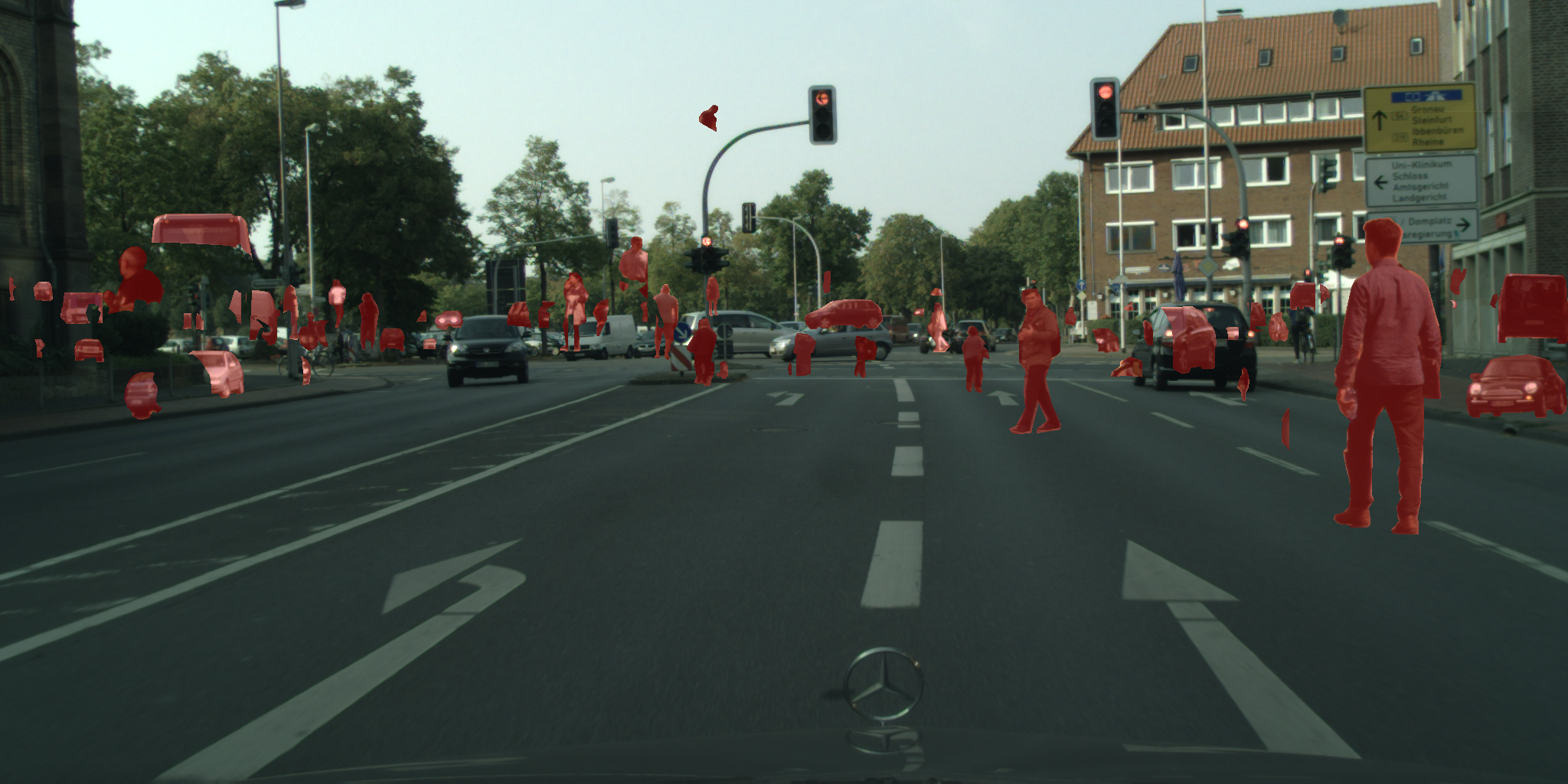}
    \caption{Input image $\textbf{x}_t$\newline \newline}
    \label{subfig:amcstest-result-subfiga}
            \end{subfigure}
    \begin{subfigure}[b]{0.19\textwidth}
                    \centering
     \includegraphics[width=\textwidth]{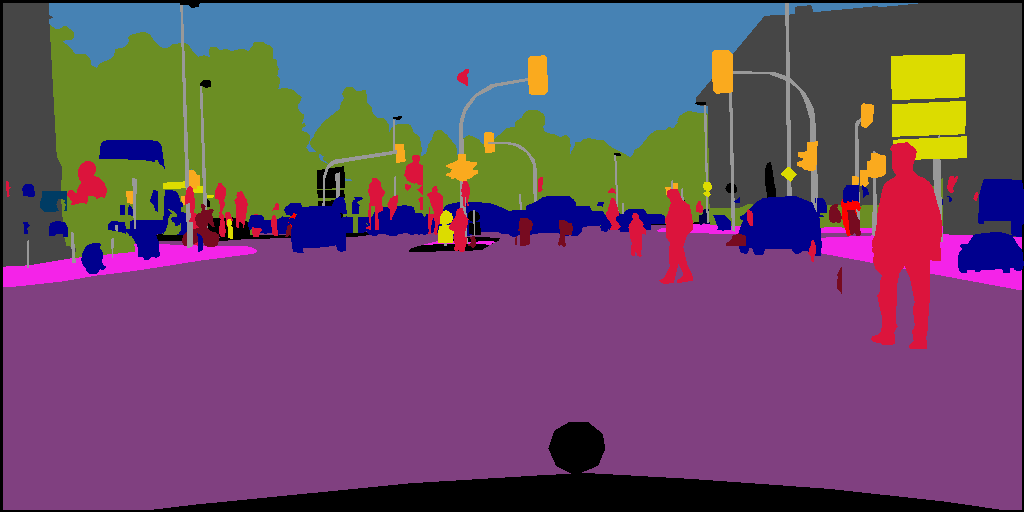}
    \caption{Ground truth semantic segmentation $\overline{\textbf{m}}_{t,f=1}$ \newline}
    \label{subfig:amcstest-result-subfigb}
            \end{subfigure}
                \begin{subfigure}[b]{0.19\textwidth}
                    \centering
     \includegraphics[width=\textwidth]{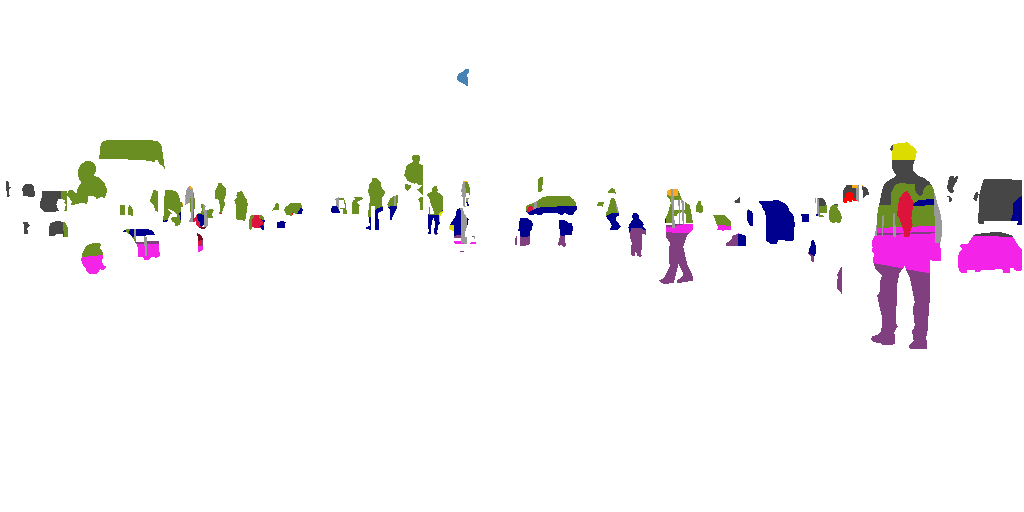}
    \caption{Ground truth amodal se\-man\-tic segmentation $\overline{\textbf{m}}_{t,f=2}$}
    \label{subfig:amcstest-result-subfigc}
            \end{subfigure}
    \begin{subfigure}[b]{0.19\textwidth}
                    \centering
     \includegraphics[width=\textwidth]{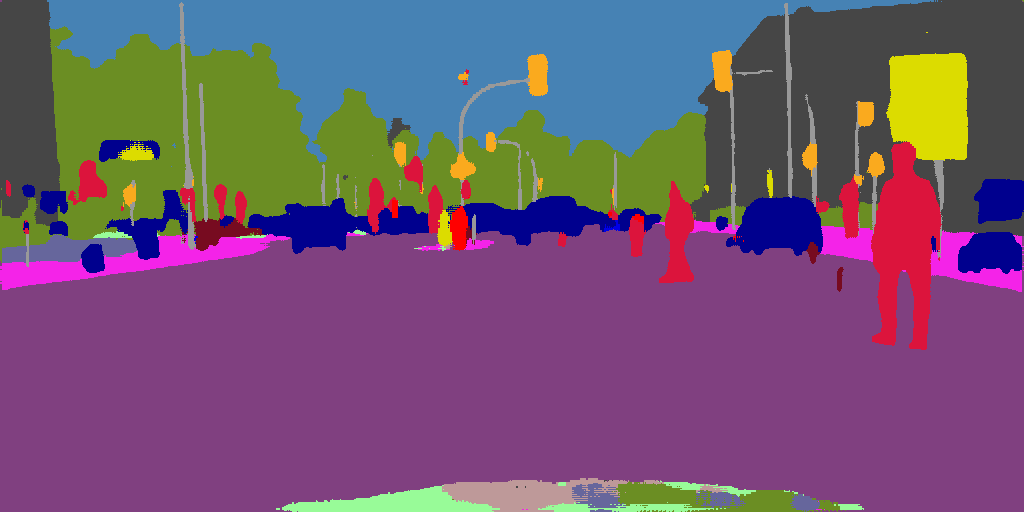}
    \caption{Predicted semantic segmentation $\textbf{m}_{t,f=1}$ \newline}
    \label{subfig:amcstest-result-subfigd}
            \end{subfigure}
                \begin{subfigure}[b]{0.19\textwidth}
                    \centering
     \includegraphics[width=\textwidth]{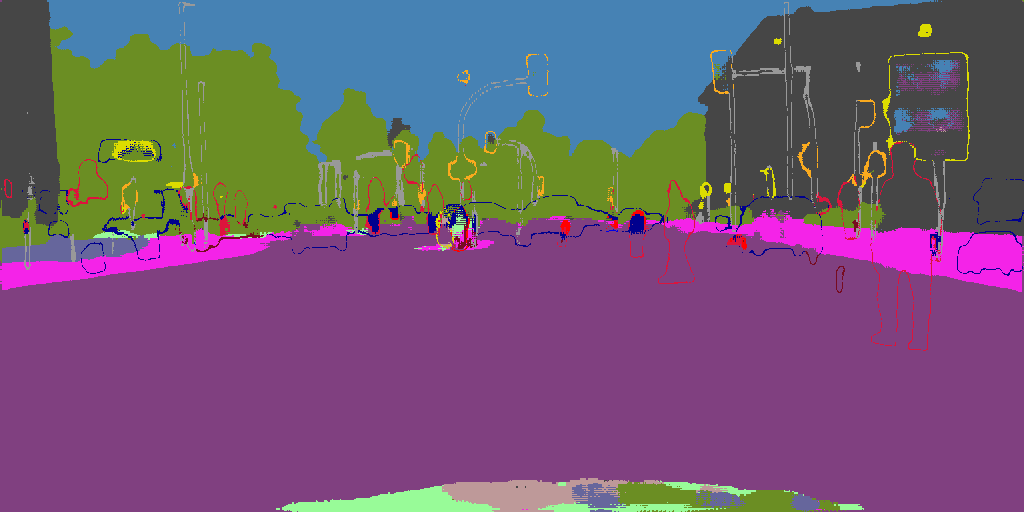}
    \caption{\textcolor{black}{Predicted amodal se\-mantic segmentation $\textbf{m}_{t,f=2}$}}
    \label{subfig:amcstest-result-subfige}
    \label{subfig:mtf2}
            \end{subfigure}
    \begin{subfigure}[b]{0.19\textwidth}
                    \centering
     \includegraphics[width=\textwidth]{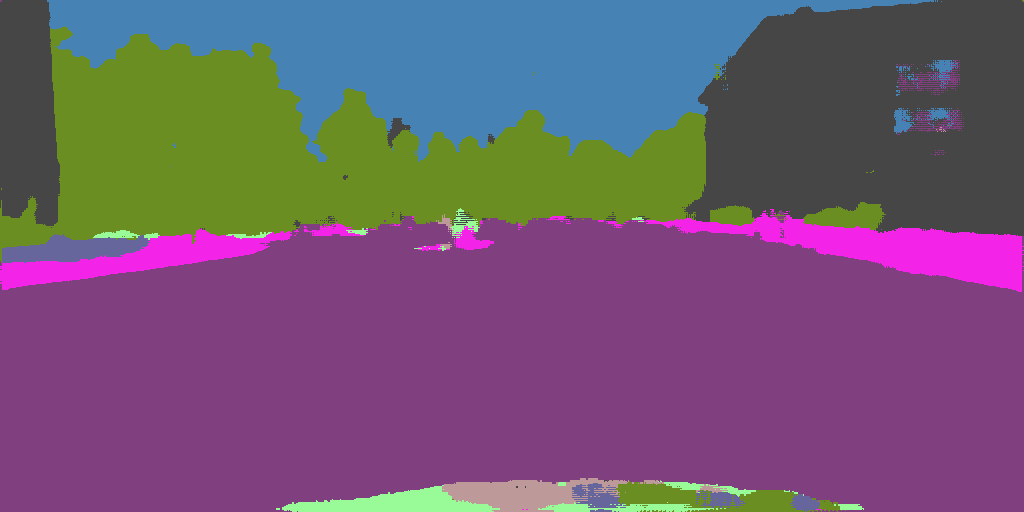}
    \caption{Predicted labels for group $\mathcal{G}_0$: $\text{argmax}(\mathbf{q}_0)$}
    \label{subfig:amcstest-result-subfigf}
            \end{subfigure}
        \begin{subfigure}[b]{0.19\textwidth}
                    \centering
     \includegraphics[width=\textwidth]{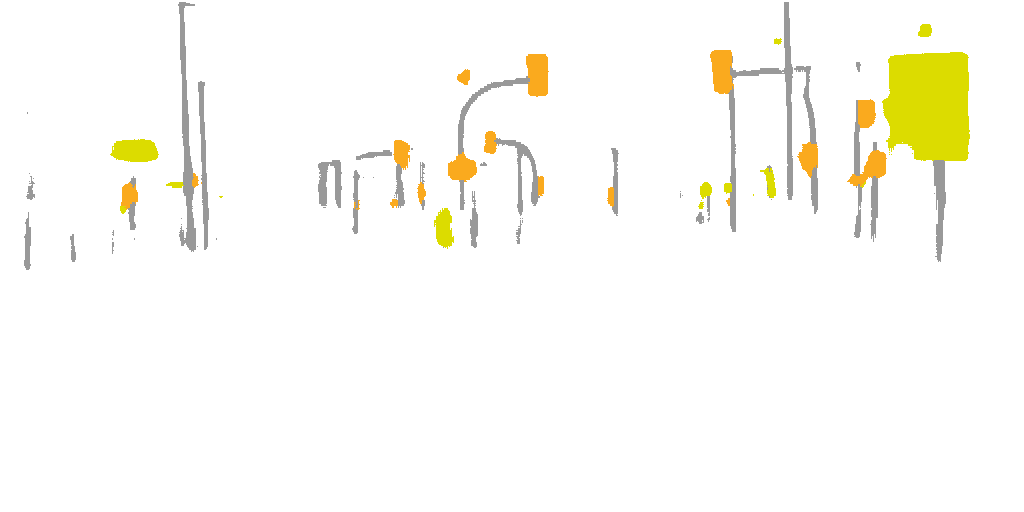}
    \caption{Predicted labels for group $\mathcal{G}_1$: $\text{argmax}(\mathbf{q}_1)$}
    \label{subfig:amcstest-result-subfigg}
            \end{subfigure}
                \begin{subfigure}[b]{0.19\textwidth}
                    \centering
     \includegraphics[width=\textwidth]{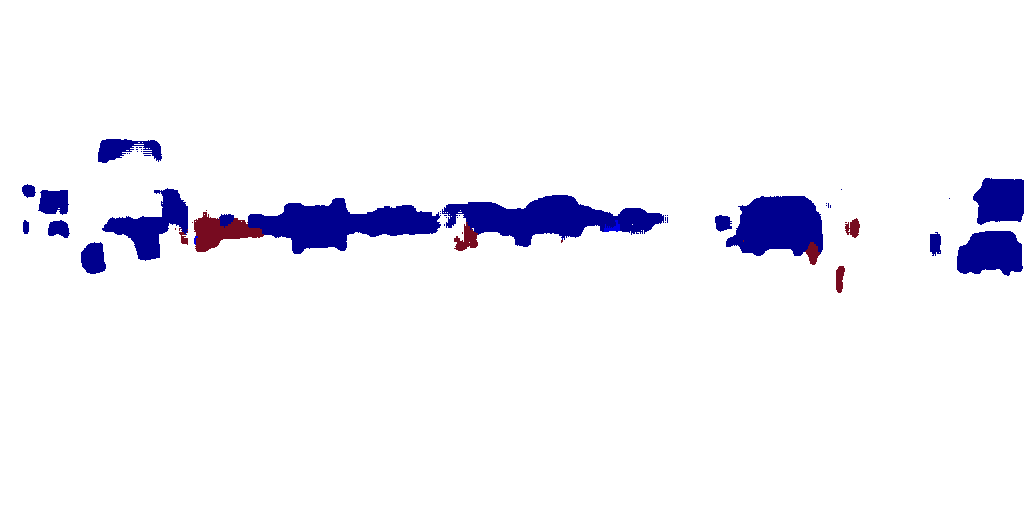}
    \caption{Predicted labels for group $\mathcal{G}_2$: $\text{argmax}(\mathbf{q}_2)$}
    \label{subfig:amcstest-result-subfigh}
            \end{subfigure}
                \begin{subfigure}[b]{0.19\textwidth}
                    \centering
     \includegraphics[width=\textwidth]{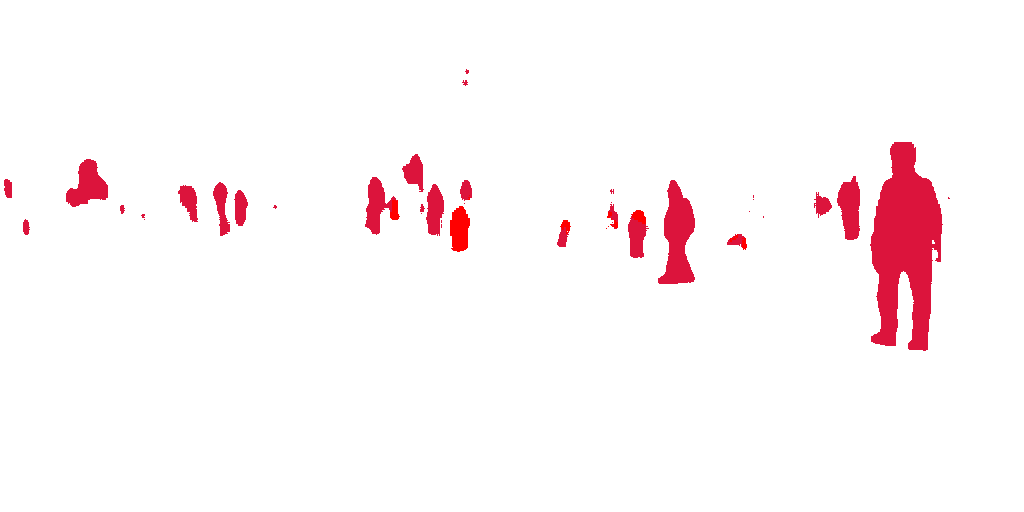}
    \caption{Predicted labels for group $\mathcal{G}_3$: $\text{argmax}(\mathbf{q}_3)$}
    \label{subfig:amcstest-result-subfigi}
            \end{subfigure}
    \caption{Qualitative results of the amodal method on an image from our Amodal Cityscapes test set.}
    \label{fig:results_on_amodalcs}
\end{figure*}

 Examples are shown in Figure~\ref{subfig:amcstest-result-subfigf}- (i). Purkait et al. \cite{Purkait2019} propose in their work a loss to take both visible and amodal predictions into account. It combines the cross-entropy loss on the visible groups with weighted cross-entropy losses for each group for visible and occluded pixels as well as the possibility that a class and group is absent in a pixel position. The details can be found in \cite{Purkait2019}. We employ this loss to train the amodal semantic segmentation method. We term this network \texttt{amERFNet} denoting the extension of the standard \texttt{ERFNet} to amodal semantic segmentation with $K$ groups.

\section{Experimental Evaluation and Discussion}

In this section, we specify the metrics needed to evaluate amodal semantic segmentation, and define the corresponding Amodal Cityscapes task (or: challenge). Then, we give the training details of the baseline method and evaluate it according to the defined challenge task.

\subsection{Metrics}
For evaluation of amodal semantic segmentation on the generated dataset, we propose the following setup with evaluation metrics based on the mean intersection over union (mIoU), which is typically used in semantic segmentation. In the case of our baseline method, we additionally report the mIoU for the underlying \texttt{ERFNet} without the amodal modifications. 
To evaluate amodal semantic segmentation methods, we first report the mIoU = mIoU$^\text{vis}$, which is calculated on the \textit{visible} parts of the images. Additionally, we report two metrics which reflect the quality of the amodal segmentation.
The first metric is the \textit{invisible} mIoU:
\begin{equation}
    \text{mIoU}^\text{inv} = \frac{1}{S} \sum\limits_{s \in \mathcal{S}}\frac{\text{TP}^\text{inv}_s}{\text{TP}^\text{inv}_s +\text{FP}^\text{inv}_s +\text{FN}^\text{inv}_s}.
\end{equation}
This metric can only be calculated on the areas of the image, where an occluding instance was pasted, since there we know about the occluded objects ($f=2$). Hence, true positives ($\text{TP}^\text{inv}_s$) per class $s \in \mathcal{S}$ are defined by images with index $t$ and pixel positions $i$, where $\overline{m}_{t,i,2}=m_{t,i,2}=s$ holds, wherein the latter denotes the network prediction. False positives ($\text{FP}^\text{inv}_s$) are defined by $\overline{m}_{t,i,2} \neq s=m_{t,i,2}$, and false negatives ($\text{FN}^\text{inv}_s$) by $\overline{m}_{t,i,2} = s \neq m_{t,i,2} $. In all cases, the pixel position $i$ is restricted to the areas where ground truth knowledge about occluded objects is available: $i \in \lbrace j \in \mathcal{I} | a_{n,j}=1, n \in \mathcal{N} \rbrace$. 
Additionally, a total quality measure is necessary to report the overall quality of amodal semantic segmentation on both the visible and occluded areas. The total mIoU according to \cite{Purkait2019} is defined as
\begin{equation}
    \text{mIoU}^\text{total} = \frac{1}{S} \sum\limits_{s \in \mathcal{S}} \frac{\text{TP}^\text{total}_s}{\text{TP}^\text{total}_s +\text{FP}^\text{total}_s +\text{FN}^\text{total}_s}.
\end{equation}
We consider a TP$_s^\text{total}$ if class $s$ is present (visible or occluded) in a pixel position $i$ both in the ground truth and in the prediction, i.e., if $\left( \overline{m}_{t,i,1}=s = m_{t,i,1} \right) \vee \left( \overline{m}_{t,i,2}=s = m_{t,i,2} \right)$. For an FP$_s^\text{total}$, we have $\left( \overline{m}_{t,i,1} \neq s=m_{t,i,1} \right) \vee \left( \overline{m}_{t,i,2} \neq s=m_{t,i,2} \right)$, while for an FN$_s^\text{total}$, $\left( \overline{m}_{t,i,1} = s \neq m_{t,i,1}\right) \vee \left( \overline{m}_{t,i,2} = s \neq m_{t,i,2} \right)$.





\begin{table}[t!]
    \centering
    \begin{tabular}{c|c|c|c|c|c}
    \hline
        Method & $K$ & Dataset & $\text{mIoU}$ & $\text{mIoU}^\text{inv} $ & $\text{mIoU}^\text{total} $  \\
         \hline \hline
                 \texttt{ERFNet} & - & \stz $\mathcal{D}_\text{amCS}^\text{test}$ &\textcolor{black}{$62.99\%$} & \textcolor{black}{$5.00\%$} & \textcolor{black}{$55.82 \%$} \\
        \texttt{amERFNet} &$3$ & \stz $\mathcal{D}_\text{amCS}^\text{test}$ &\textcolor{black}{$61.93 \%$} & \textcolor{black}{$18.52\%$} & \textcolor{black}{$58.03 \%$} \\
        \texttt{amERFNet}& $4$ & \stz \stzdown $\mathcal{D}_\text{amCS}^\text{test}$ &\textcolor{black}{$62.76 \%$} & \textcolor{black}{$23.60\%$} & \textcolor{black}{$59.12\%$} \\
         \texttt{ERFNet} \stz & - & \stz $\mathcal{D}_\text{CS}^\text{val}$ &\textcolor{black}{$67.97 \%$} & $\ast$ & $\ast$ \\
        \texttt{amERFNet}& $3$ & \stz $\mathcal{D}_\text{CS}^\text{val}$ &\textcolor{black}{$66.71\%$} & $\ast$ & $\ast$ \\
        \texttt{amERFNet} & $4$ & \stz $\mathcal{D}_\text{CS}^\text{val}$ &\textcolor{black}{$68.53\%$} & $\ast$ & $\ast$ \\
        \hline
    \end{tabular}
    \caption{Performance of \texttt{ERFNet} vs.\ \texttt{amERFNet} (Amodal Cityscapes Challenge baseline) on the original and the Amodal Cityscapes dataset. Fields marked with $\ast$ cannot be calculated due to missing amodal ground truths.}
    \label{tab:results}
\end{table}

\subsection{Implementation and Training Details}

The network is trained on the Amodal Cityscapes training set $\mathcal{D}_\text{amCS}^\text{train}$. We monitor the training using the validation split $\mathcal{D}_\text{amCS}^\text{val}$. The exact split can be found on github.
We use the Adam optimizer with an initial learning rate of $0.01$ and exponential decay to train the amodal semantic segmentation for 120 epochs. We select the best performing model on the validation set for evaluation on the test set.

\subsection{Segmentation Results and Discussion}
We show the results of our proposed new Amodal Cityscapes baseline method in Table \ref{tab:results}. 
The first three rows report the results of the standard semantic segmentation and the amodal \texttt{ERFNet} on the Amodal Cityscapes test set $\mathcal{D}_\text{amCS}^\text{test}$.
We see that all three methods achieve similar mIoU values, with the \texttt{ERFNet} performing only slightly better ($62.99\%$) than the amodal methods ($61.93\%/62.76\%$). Accordingly, our results show that the mIoU does not suffer considerably from learning to additionally predict the amodal labels. The \texttt{ERFNet} on the other hand, is not at all able to predict amodal labels.
The corresponding poor mIoU$^\text{inv}=5.00\%$ is not equal to zero due to predicted labels that by chance coincide with the amodal ground truth. This leads then to an mIoU$^\text{total}$ of $55.82\%$. Our method ($K=3$) provides a clearly better-than-chance mIoU$^\text{inv}$ of $18.52\%$, which significantly increases further for $K=4$ ($23.60\%$). The mIoU$^\text{total}$ of the \texttt{ERFNet} is outperformed by our amodal semantic segmentation methods by more than $2\%$ absolute, which can be dedicated to the increased mIoU$^\text{inv}$ for both \texttt{amERFNet} methods. Note that the large total amount of visible scenes compared to the small amount of occluded areas with available ground truth compensates the low mIoU$^\text{inv}$ in the mIoU$^\text{total}$, hence the mIoU$^\text{total}$ discrepancy between the amodal and standard semantic segmentation does not mirror the drastic improvements in mIoU$^\text{inv}$. 

For the method with the highest mIoU$^\text{total}$ (\texttt{amERFNet} with $K=4$), we visualize an example result in Figure~\ref{fig:results_on_amodalcs}. Here, we see the input image, the corresponding ground truth and predicted segmentation masks. Additionally, we show the predicted labels in each group in the bottom row. We observe that especially the static classes of $\mathcal{G}_0$ can be recovered well, and the predicted amodal semantic segmentation mask $\mathbf{m}_{t,f=2}$ in Figure~\ref{subfig:mtf2} provides a clearly better-than-chance prediction of the occluded areas.   

\begin{figure}[t!]
\vspace*{2pt}
    \centering
    \begin{subfigure}[b]{0.49\textwidth}
                    \centering
        \includegraphics[width=0.475\linewidth,height=2.07cm]{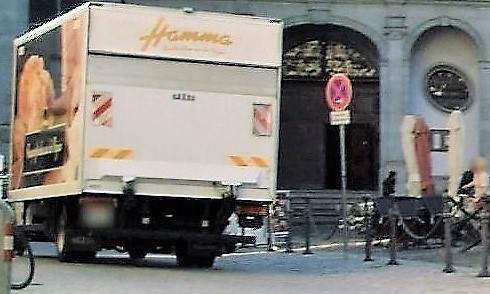}
       \includegraphics[width=0.475\linewidth]{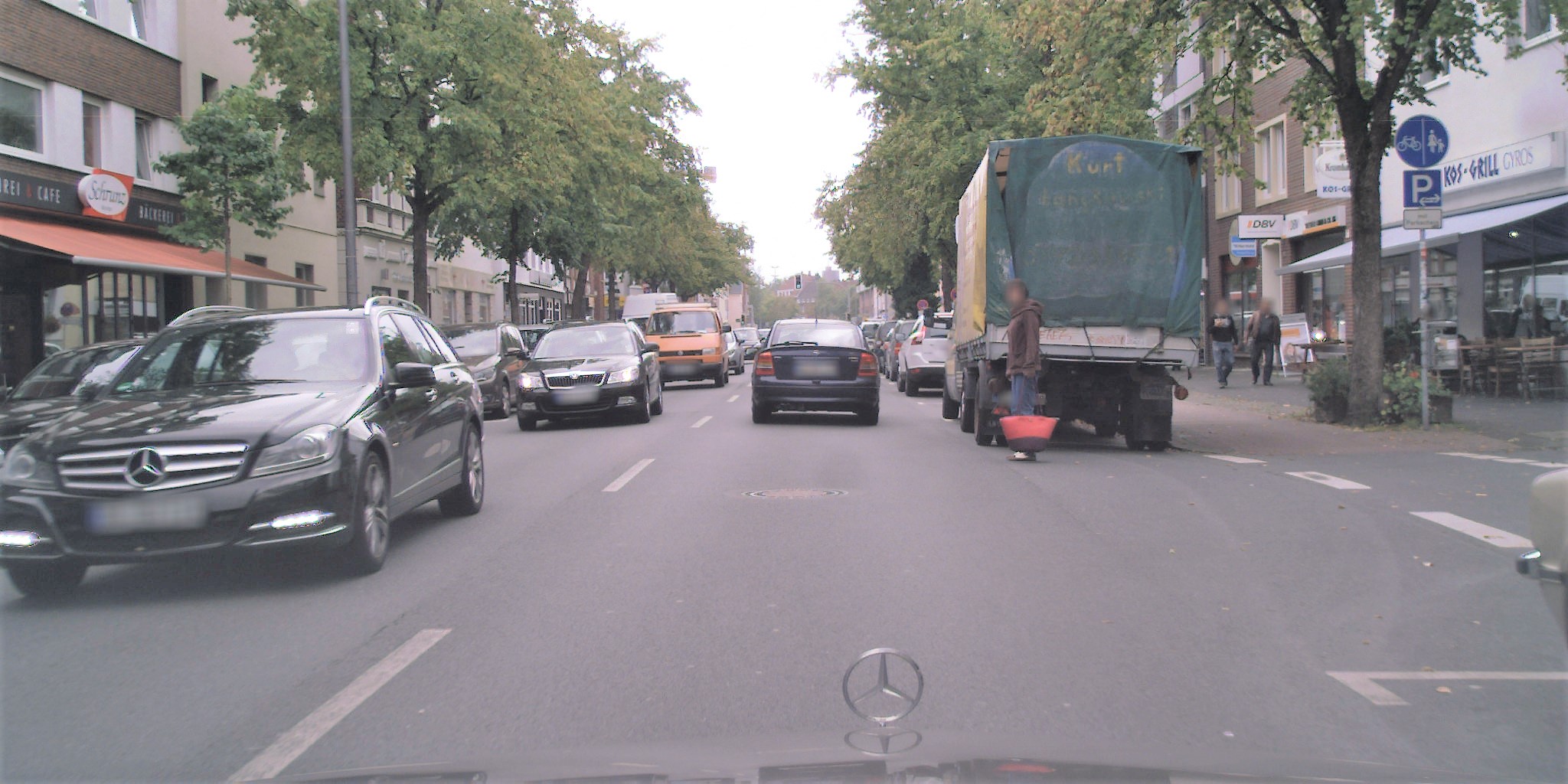}
        \caption{Input image $\textbf{x}_t$}
        \label{subfig:csval-examples-images}
    \end{subfigure}
        \begin{subfigure}[b]{0.49\textwidth}
                    \centering
        \includegraphics[width=0.475\textwidth,height=2.07cm]{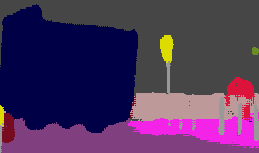}
        \includegraphics[width=0.475\textwidth]{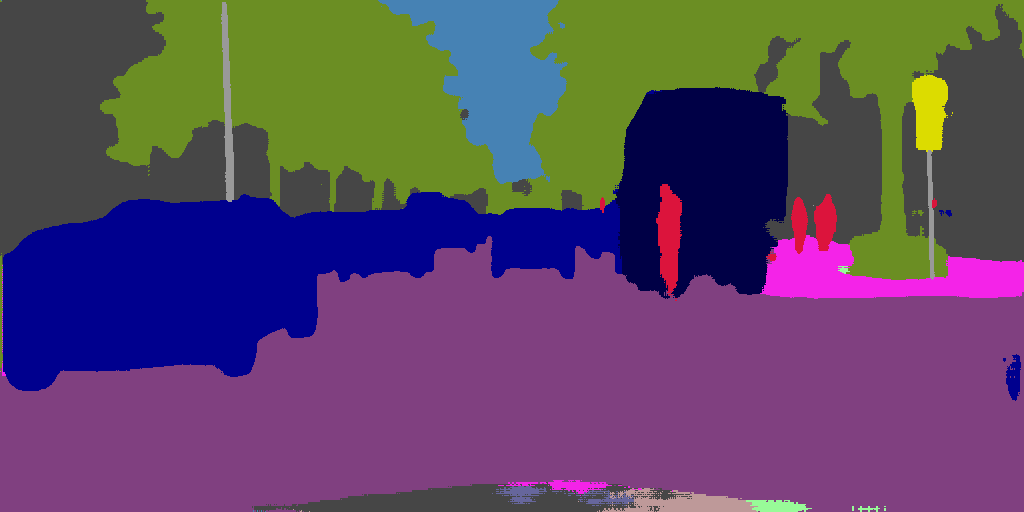}
        \caption{Se\-mantic segmentation $\textbf{m}_{t,f=1}$}
        \label{subfig:csval-examples-visible}
    \end{subfigure}
            \begin{subfigure}[b]{0.49\textwidth}
                    \centering
        \includegraphics[width=0.475\textwidth,height=2.07cm]{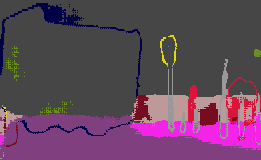}
        \includegraphics[width=0.475\textwidth]{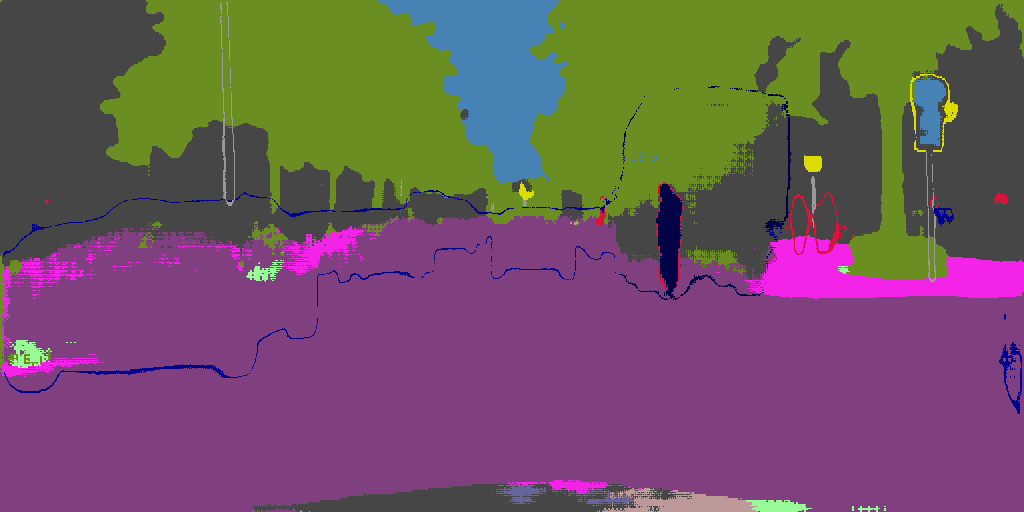}
        \caption{Amodal se\-mantic segmentation $\textbf{m}_{t,f=2}$}
        \label{subfig:csval-examples-invisible}
    \end{subfigure}
    \caption{Example results (image crops) of the $\texttt{amERFNet}$ on the standard Cityscapes validation set $\mathcal{D}_\text{CS}^\text{val}$.}
    \label{fig:amodal-results-csval}
\end{figure}

The last three rows of Table \ref{tab:results} report the results of the \texttt{ERFNet} and the amodal \texttt{ERFNet} on the Cityscapes validation set $\mathcal{D}_\text{CS}^\text{val}$. As no amodal labels are available for $\mathcal{D}_\text{CS}^\text{val}$, we are unable to report on $\text{mIoU}^\text{inv}$ and $\text{mIoU}^\text{total}$.

The semantic segmentation of the \texttt{ERFNet} and the \texttt{amERFNet} with $K=4$ achieve similar (visible) mIoU values ($67.97\% \slash 68.53\%$) on $\mathcal{D}_\text{CS}^\text{val}$. The \texttt{amERFNet} with $K=3$ lacks a bit behind ($66.71\%$). Although no amodal ground truth is available on $\mathcal{D}_\text{CS}^\text{val}$, in Figure~\ref{fig:amodal-results-csval}, we show exemplary visible (Figure~\ref{subfig:csval-examples-visible}) and amodal segmentations (Figure~\ref{subfig:csval-examples-invisible}) using the \texttt{amERFNet} with $K=4$ of two images of $\mathcal{D}_\text{CS}^\text{val}$. In both cases we see a plausible hallucination mostly of static classes (group $\mathcal{G}_0$) in the amodal semantic segmentation masks $m_{t,f=2}$ in Figure~\ref{subfig:csval-examples-invisible}. The still visible object contours are likely due to label ambiguity and the blurring of instance contours in the training data. A possible remedy could be image processing algorithms such as erosion and dilation. Considering the left image in Figure~\ref{subfig:csval-examples-images}, and the corresponding predicted masks in Figures \ref{subfig:csval-examples-visible} and \ref{subfig:csval-examples-invisible}, on the right side of the left image in Figure~\ref{subfig:csval-examples-images} persons are partially occluded by the light brown objects visualized in the semantic segmentation (left mask of Figure~\ref{subfig:csval-examples-visible}). The occluded shape of those persons is anticipated (in red) in the left mask of Figure~\ref{subfig:csval-examples-invisible}. The upper bodies are visible, and thus shown in red in the semantic segmentation (Figure~\ref{subfig:csval-examples-visible}, right side of the left mask). One person close to the door of the building in the left image of Figure~\ref{subfig:csval-examples-images}, is not detected in the visible mask (Figure~\ref{subfig:csval-examples-visible}), but its entire shape is hallucinated in the invisible mask (left mask of Figure~\ref{subfig:csval-examples-invisible}). The background of the truck consisting of classes belonging to the static group has been recovered reasonably (left mask of Figure~\ref{subfig:csval-examples-invisible}).
Concerning now the image and masks on the right of Figure~\ref{fig:amodal-results-csval}, on the right of Figure~\ref{subfig:csval-examples-visible} a person is standing in front of the truck. Figure~\ref{subfig:csval-examples-invisible} on the right shows that the predicted amodal semantic segmentation mask is able to hallucinate the truck behind that person, while in the other non-occluded areas of the truck even the static classes in the background are correctly predicted in the amodal semantic segmentation $\mathbf{m}_{t,f=2}$.

\section{Conclusions}

In this paper, we consider the task of amodal semantic segmentation of single images. As amodal perception is a critical ability of intelligent vehicles, reliably seeing behind occlusions is important. To facilitate training of amodal semantic segmentation methods, we propose a generic way to create copy-paste amodal datasets with plausible occluder locations and sizes and, exemplary, create an \textit{amodal Cityscapes} dataset. Additionally, we evaluate a baseline for amodal semantic segmentation on the generated dataset. Both the dataset and the baseline shall serve as a challenge to develop and evaluate methods for amodal semantic segmentation to improve perception of intelligent vehicles. \textit{Our aim is to provide a playground to advance amodal semantic segmentation on automotive tasks.} To this end, the amodal Cityscapes dataset forms a data basis to develop and train methods, as well as to evaluate them. To ensure comparability between methods, we publish the file lists\footnote{\textcolor{black}{\url{https://github.com/ifnspaml/AmodalCityscapes}}} to allow to re-create our version of the amodal Cityscapes dataset. Also the scripts for dataset re-generation and for the proposed Amodal Cityscapes Challenge baseline method will be made available on github. 
Amodal semantic segmentation methods can then be evaluated on the Amodal Cityscapes test dataset as was done for the challenge baseline method. While the amodal \texttt{ERFNet} improves upon the standard \texttt{ERFNet} by providing a better-than-chance invisible mIoU (mIoU$^\text{inv}$), and hence an improved total mIoU (mIoU$^\text{total}$), there is still plenty of room for advances towards high-performing amodal semantic segmentation methods. 
With an evaluation that focuses not only on correctly predicting behind occlusions, but also the mIoU$^\text{total}$, we ensure that performance on the standard visible dataset does not degrade. 






\section*{Acknowledgment}
    This work results from the project KI Data Tooling (19A20001M) funded by the German Federal Ministry for Economic Affairs and Climate Action (BMWK). The authors would like to thank Hanh Thi My Nguyen for her contributions to this work. 





{\small
\bibliographystyle{IEEEbib}
\bibliography{bibtex}
}

\end{document}